\documentclass{article}

\usepackage{arxiv}

\usepackage[utf8]{inputenc} 
\usepackage[T1]{fontenc}    
\usepackage{hyperref}       
\usepackage{url}            
\usepackage{booktabs}       
\usepackage{amsfonts}       
\usepackage{nicefrac}       
\usepackage{microtype}      
\usepackage{lipsum}
\usepackage{graphicx}
\usepackage{amsmath,amsfonts}
\usepackage{algorithmic}
\usepackage{algorithm}
\usepackage{array}
\usepackage[caption=false,font=normalsize,labelfont=sf,textfont=sf]{subfig}
\usepackage{textcomp}
\usepackage{stfloats}
\usepackage{url}
\usepackage{verbatim}
\usepackage{graphicx}
\usepackage{cite}
\usepackage{xcolor}
\usepackage{amsmath}
\usepackage{makecell}
\usepackage{multicol}
\usepackage{multirow}
\usepackage{booktabs}
\usepackage{siunitx}
\usepackage{gensymb}
\usepackage{soul, color, xcolor}
\usepackage[utf8]{inputenc}
\usepackage[english]{babel}
\usepackage{threeparttable}
\usepackage{graphicx} 
\usepackage{amsmath}
\usepackage{amssymb}
\usepackage{tablefootnote}
\usepackage{stfloats}
\usepackage{threeparttable}
\usepackage{color,soul}
\graphicspath{ {./images/} }
\usepackage{textcomp}
\usepackage{microtype}
\usepackage{siunitx}

\title{Enhancing the Performance of a Biomimetic Robotic Elbow-and-Forearm System Through Bionics-Inspired Optimization}

\author{
Haosen Yang \\
  The Department of Mechanical, Aerospace and Civil Engineering\\
  University of Manchester\\
  Manchester, M13 9PL, UK \\
  \texttt{haosen.yang@postgrad.manchester.ac.uk} \\
   \And
 Guowu Wei \\
  School of Science, Engineering and Environment\\
  University of Salford\\
  Salford, M5 4WT, UK \\
  \texttt{g.wei@salford.ac.uk} \\
  \And
Lei Ren \\
  The Key Laboratory of Bionic Engineering, Ministry of Education\\
  Jilin University\\
  Changchun 130025, China \\
  \texttt{lren@jlu.edu.cn} \\
}

\begin{document}
\maketitle
\begin{abstract}
This paper delineates the formulation and verification of an innovative robotic forearm and elbow design, mirroring the intricate biomechanics of human skeletal and ligament systems. Conventional robotic models often undervalue the substantial function of soft tissues, leading to a compromise between compactness, safety, stability, and range of motion. In contrast, this study proposes a holistic replication of biological joints, encompassing bones, cartilage, ligaments, and tendons, culminating in a biomimetic robot. The research underscores the compact and stable structure of the human forearm, attributable to a tri-bone framework and diverse soft tissues. The methodology involves exhaustive examinations of human anatomy, succeeded by a theoretical exploration of the contribution of soft tissues to the stability of the prototype. The evaluation results unveil remarkable parallels between the range of motion of the robotic joints and their human counterparts. The robotic elbow emulates 98.8\% of the biological elbow's range of motion, with high torque capacities of 11.25 Nm (extension) and 24 Nm (flexion). Similarly, the robotic forearm achieves 58.6\% of the human forearm's rotational range, generating substantial output torques of 14 Nm (pronation) and 7.8 Nm (supination). Moreover, the prototype exhibits significant load-bearing abilities, resisting a 5kg dumbbell load without substantial displacement. It demonstrates a payload capacity exceeding 4kg and rapid action capabilities, such as lifting a 2kg dumbbell at a speed of 0.74Hz and striking a ping-pong ball at an end-effector speed of 3.2m/s. This research underscores that a detailed anatomical study can address existing robotic design obstacles, optimize performance and anthropomorphic resemblance, and reaffirm traditional anatomical principles.
\end{abstract}
\section{Introduction}

In recent years, significant advancements in the robotics field have focused on developing and controlling humanoid robots for integration into daily life. These robots are designed to interact with humans and perform a variety of tasks. One envisioned scenario involves physical collaboration between humans and robots, which has long captivated the scientific community. Human-centred and ergonomic design are crucial aspects of engineering, and when humans interact with robots, safety and system efficiency are the primary considerations. The pursuit of a biomimetic appearance resembling the human body is also a key direction of effort in this field. Numerous studies have focused on developing control architectures for ergonomic physical human-robot interaction\cite{hyon2007full,zacharaki2020safety}. However, the hardware design of humanoid robots has rarely been considered for optimization in collaborative actions and is often assumed as a given. This paper contributes to the development of optimal biomimetic robotic elbow and forearm designs, grounded in human anatomical structures, to enhance performance and ergonomics in human-robot collaborative tasks.

The elbow and forearm are crucial components of the upper limb. In traditional robotic arm designs, the forearm typically features a geared motor directly connected to the forearm output rotation, with two rotating joints in series to mimic elbow flexion/extension and forearm rotation\cite{grebenstein2011dlr,paik2012development}. These designs offer several advantages, such as a large range of motion\cite{ogura2006development,englsberger2014overview,yu2014design}. Ultra-powerful motors can generate considerable torque by increasing motor and limb size\cite{walker1994modular,borst2009rollin}. They can achieve exceptional strength and ultra-high accuracy using materials like stainless steel, aluminium, titanium, hinged joints, high-precision gearboxes, and advanced manufacturing technology. Moreover, these designs can simplify the design, manufacturing, and maintenance processes. However, balancing compactness and high output performance can be challenging since the motor needs to be installed near the joint for optimal efficiency. For example, using a small motor for forearm rotation may result in insufficient output torque, while an excessively large motor can lead to a bulky forearm, taking up space within the forearm structure and complicating the installation of muscles responsible for hand joint movements when using remote tendon control. On the other hand, achieving compactness in the forearm often requires local control of hand actuation, with all hand actuators located inside the hand, making it difficult to generate larger output torque at finger joints. Moreover, as human-robot interaction requirements increase, the rigidity and power of such robotic systems can pose safety risks during interactions. Additionally, many robots lack the natural, human-like aesthetics needed for comfortable interaction.

The structure of the human forearm and elbow joint has several advantages. Firstly, its compact nature accommodates robust muscles within a small forearm diameter, enabling intricate and precise hand and wrist movements, hence contributing to manual dexterity. Secondly, its stability, reflected in mobile yet resilient joints that resist easy dislocation, permits the handling of heavy tasks without elbow and forearm damage, thereby facilitating significant load-bearing capacity. The third attribute pertains to safety and compliance. Unlike conventional rigid joints, human joints exhibit damping and elastic properties, thereby offering variable joint stiffness. A salient feature of biological joints is their capacity to dislocate under extreme external forces and their inherent self-recovery mechanisms. This process echoes the actions of an orthopaedic surgeon in treating a dislocated human joint. {The human joints have the capability to heal over time, traditional robots, on the other hand, require external intervention for repairs. For enhanced operator safety, robotic joints designed to permit controlled dislocation, and subsequently be straightforward "reset", facilitate rapid repairs and swift return to operational status, all while safeguarding the user.} The potential of these characteristics to enhance robotics and automation systems warrants further exploration. Incorporating this feature into a robotic arm can enhance operator safety, appropriate in a human-robot interaction environment. Consequently, many researchers have developed biomimetic designs that emulate the human structure \cite{hosoda2012anthropomorphic,seo2019human,diamond2012anthropomimetic,mizuuchi2006development,sodeyama2008designs,mouthuy2022humanoid,trendel2018cardsflow}. Some of these designs have adopted a tendon-driven approach akin to the biological arm, leveraging the physical properties of tendons to mimic the inherent compliance and dynamics of musculoskeletal characteristics. Additionally, some have achieved a more closely resembling appearance to a biological arm. While these designs successfully replicate basic human forearm and elbow functionality, they often only partially represent human anatomy, particularly in terms of soft tissues. Insufficient soft tissue representation may result in structural stability issues, such as lateral forearm stability. Refined soft tissue representation can improve load-carrying ability, impedance, and compliance, and provide flexible limitations when the joint reaches its extreme position. Incorporating soft tissues allows the joint to recover to a certain extent when dislocated by extreme external forces, significantly enhancing the safety of human-robot interaction. Soft tissues also play a crucial role in introducing damping to the entire system, effectively mitigating oscillations that may occur during mechanical motion.

This study investigates the human forearm and elbow to understand how the three bones (humerus, ulna and radius) achieve a wide range of motion while maintaining axial and lateral load-carrying ability, compactness, and stability. This insight is then applied to the design of a biomimetic robot. Furthermore, a biomimetic actuation method was implemented in the robot to explore whether adopting a human-like actuation scheme could enhance joint output while preserving compactness. The research explores the potential advantages of this approach from an academic perspective to determine whether employing these structures can optimize the biomimetic robot's design and address inherent issues. 

\section{Related work}

\subsection{Existing robotic arm designs}

In traditional robotic arms, the elbow allows flexion/extension with a split forearm connected to the elbow for forearm rotation\cite{grebenstein2011dlr,paik2012development,ogura2006development,englsberger2014overview,yu2014design,walker1994modular,borst2009rollin}. This design paradigm has maintained preeminence in the realm of robotic arms, largely owing to its straightforward, efficient architecture that streamlines design, manufacturing, and maintenance procedures, in addition to facilitating the implementation of control algorithms. These designs often employ rigid components such as bearings and shafts to achieve a wide range of motion and stabilize the joints. In contrast, the human elbow relies on a three-bone structure, wherein the radius rotates around the ulna to accomplish forearm rotation, and both the radius and ulna rotate around the humerus to achieve elbow flexion/extension. This distinctive configuration attains two rotational joints within a compact structure, adeptly balancing mobility and stability without the dependence on shafts.

Drawing from the human skeletal system, several research groups have proposed robotic elbow and forearm mechanisms reflecting the three-bone structure of the human arm \cite{seo2019human,seo2016development,hosoda2012anthropomorphic}. These solutions, employing conventional hinge and ball-and-socket articulations, mirror the human forearm structure, facilitating the rotation of the radius around the ulna. Despite their bio-inspired designs, these systems predominantly resort to rigid architectures to imitate articulated joints, thus achieving humanoid joint motions. While they successfully address certain limitations of conventional designs, such as compactness and mobility, with \cite{seo2019human} even simulating human ligaments for enhanced safety, they often neglect to thoroughly investigate or exploit the human body's innate structural advantages. These designs generally provide a simplified representation of bodily joint functions, mostly overlooking the contributions of the human body's soft tissues. Their exclusion may lead to stability concerns or increased joint friction under substantial loading. Moreover, safety remains a substantial challenge in human-robot interaction scenarios. Our research delves into the intricacies of human joint structures, investigating their inherent mechanical properties, and utilizes this knowledge to propose an innovative elbow and forearm design addressing the aforementioned challenges. We initiate our exploration with an introduction to the human elbow and forearm's anatomical structure.

\subsection{Anatomy study of biological elbow and forearm}

\begin{figure}[htb]
\centerline{\includegraphics[width=0.6\textwidth]{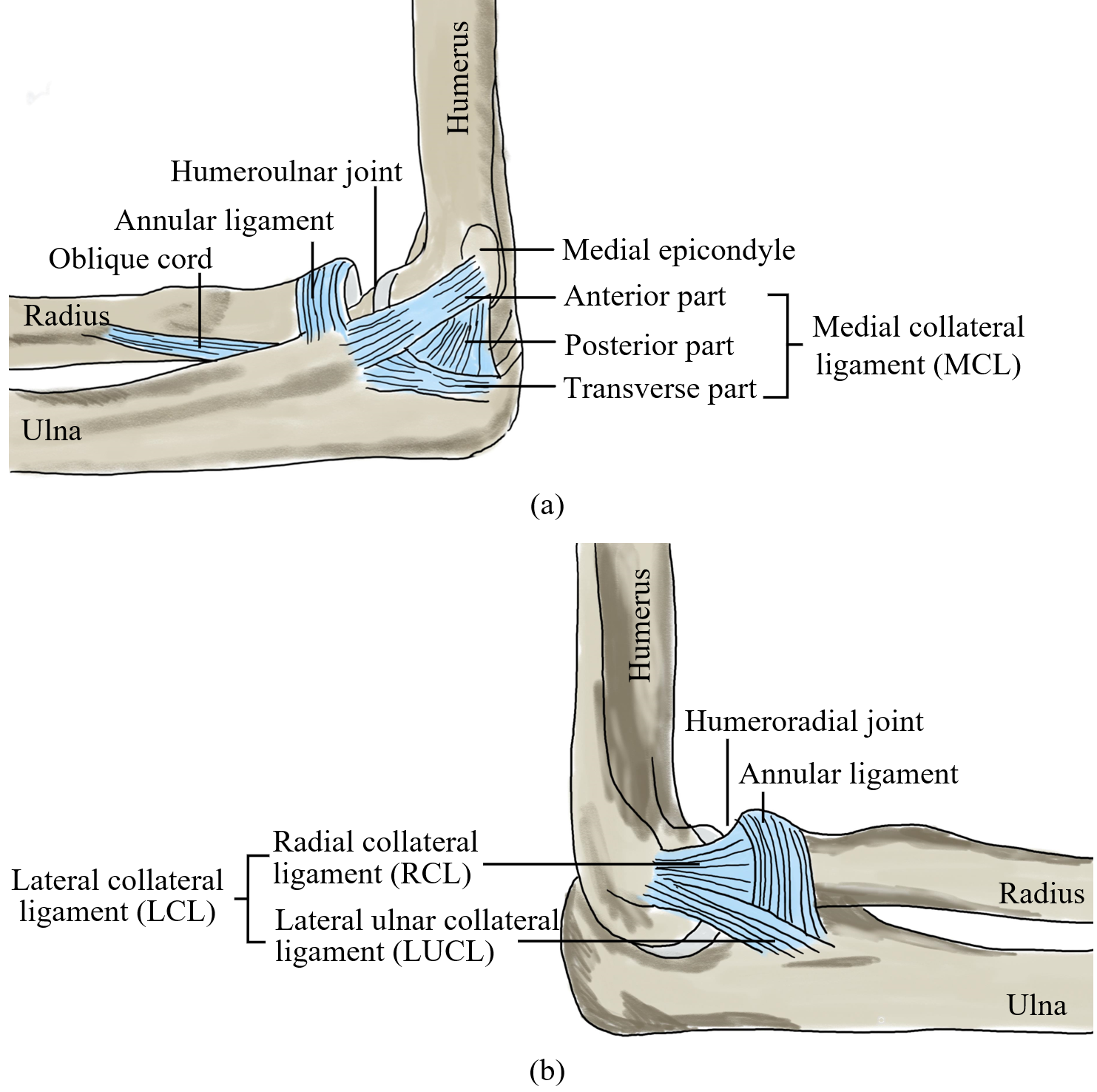}}
\caption{Bones and soft tissues in the elbow joint: (a) The medial collateral ligament (MCL); (b) The lateral collateral ligament (LCL) and the annular ligament\cite{lockard2006clinical}.}
\label{fig8.1}
\end{figure}

\begin{figure}[htb]
\centerline{\includegraphics[width=0.75\textwidth]{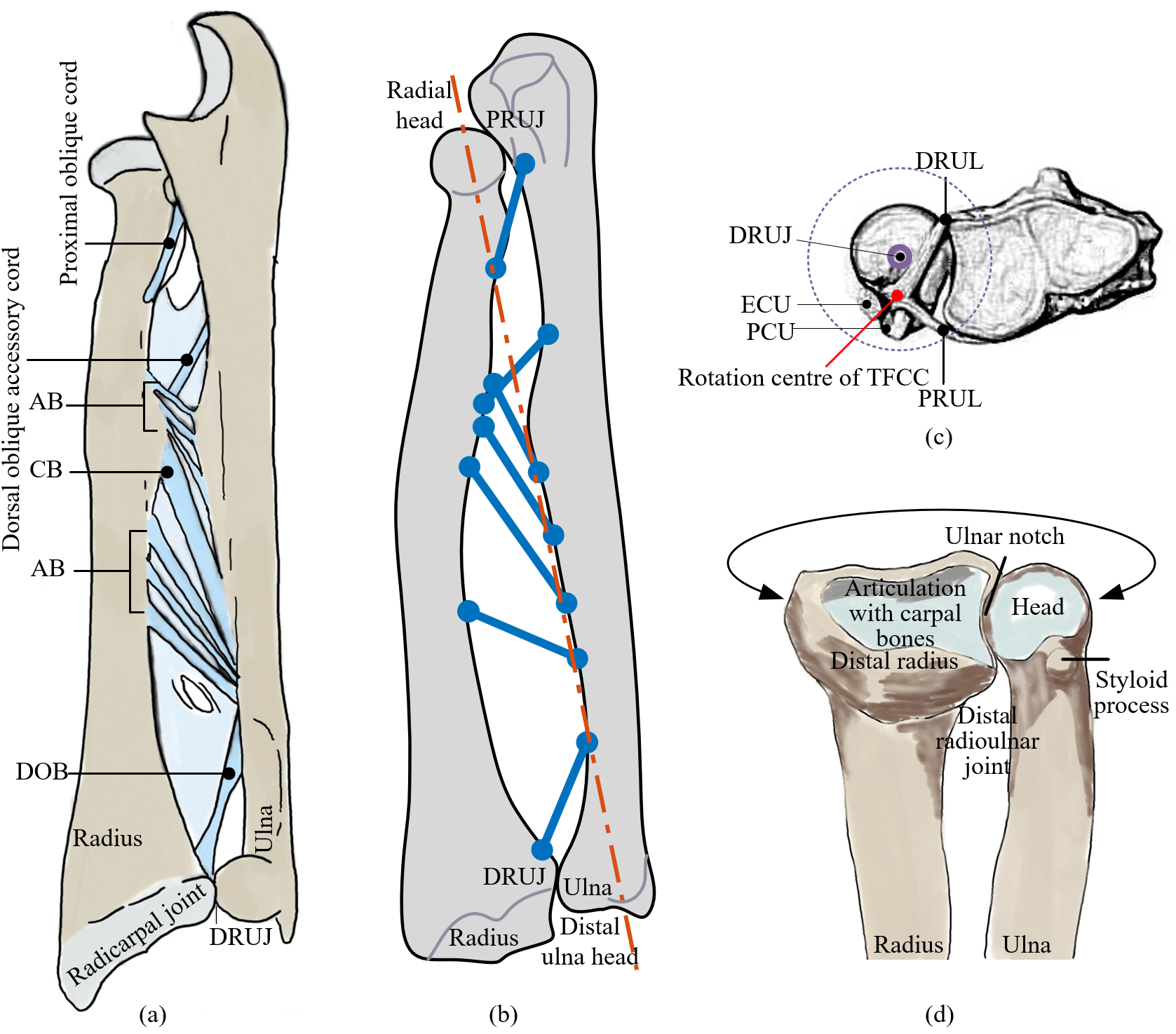}}
\caption{(a) Schematic structure of the interosseous membrane (IOM)\cite{noda2009interosseous}; (b) The insertion points of different bundles of the IOM on the ulna and radius are on the forearm rotation axis; (c) TFCC structure in the distal radioulnar joint\cite{schuenke2020general}; (d) Distal radioulnar joint (DRUJ).}
\label{fig8.3}
\end{figure}

The elbow joint is a vital component of the upper extremity, serving two primary functions in the human body. Firstly, it operates as a hinge joint, facilitating forearm flexion/extension around the humerus, which is essential for activities such as feeding, reaching, throwing, and personal hygiene. Secondly, it functions as a rotational joint in conjunction with the two radioulnar joints, enabling forearm supination/pronation, which can occur independently or simultaneously with elbow flexion/extension. Pronation and supination of the forearm enable the hand to generate rotating torque, allowing it to perform tasks such as screwing. Given its ability to operate efficiently in narrow spaces and generate omnidirectional torque. This mechanism is achieved through the rotation of the radius around the fixed ulna.

The elbow comprises two individual joints: the humeroulnar joint (Fig.\ref{fig8.1}(a)), located between the humerus and ulna, often considered a hinge joint in robotic designs. The typical range of motion for flexion/extension at the humeroulnar joint spans from 0\degree to 146\degree. The second joint, the humeroradial joint (Fig.\ref{fig8.1}(b)), can be considered as a ball-and-socket joint, is situated between the humerus and radius, consisting of flexion/extension and rotation motions.

The forearm encompasses two joints: the proximal radioulnar joint (PRUJ), positioned between the proximal ends of the ulna and radius (Fig.\ref{fig8.3}(b)), and the distal radioulnar joint (DRUJ), located at the distal ends of the ulna and radius (Fig.\ref{fig8.3}(d)). Both PRUJ and DRUJ facilitate pronation and supination of the forearm, occurring around an axis defined by a line extending from the centre of the radial head's fovea to the distal ulna head\cite{lockard2006clinical}, is represented by red line in Fig.\ref{fig8.3}(b). Pronation and supination are simple motions involving the radial head pivoting on the ulna, while the distal end of the radius glides around a stationary ulna.

Primary stability of the humeroulnar joint is ensured by two collateral ligaments of the elbow: the Medial Collateral Ligament (MCL) and the Lateral Collateral Ligament (LCL). The MCL, a critical element in maintaining elbow joint stability, comprises three primary components: anterior, posterior, and transverse bundles, as depicted in Fig.\ref{fig8.1}(a). The anterior and posterior bundles do not originate directly from the elbow rotation axis, causing variable ligament tension during flexion and extension \cite{karbach2017elbow}. Specifically, the anterior bundle experiences tension during elbow extension, while the posterior bundle is tensioned during flexion \cite{callaway1997biomechanical}. The LCL complex, another pivotal stabilizer of the elbow joint, is illustrated in Fig.\ref{fig8.1}(b). Constituting the Lateral Ulnar Collateral Ligament (LUCL), Radial Collateral Ligament (RCL), and the annular ligament, the LCL complex maintains consistent tension through the elbow's motion, given the central origin of the LUCL and RCL in relation to elbow flexion/extension \cite{nordin2001basic}. The annular ligament encapsulates the radial head and is anchored to the ulna, with the RCL's connection to the annular ligament providing further stabilization to the radial head \cite{ahmed2015management}. 

\begin{figure}[htb]
\centerline{\includegraphics[width=0.6\textwidth]{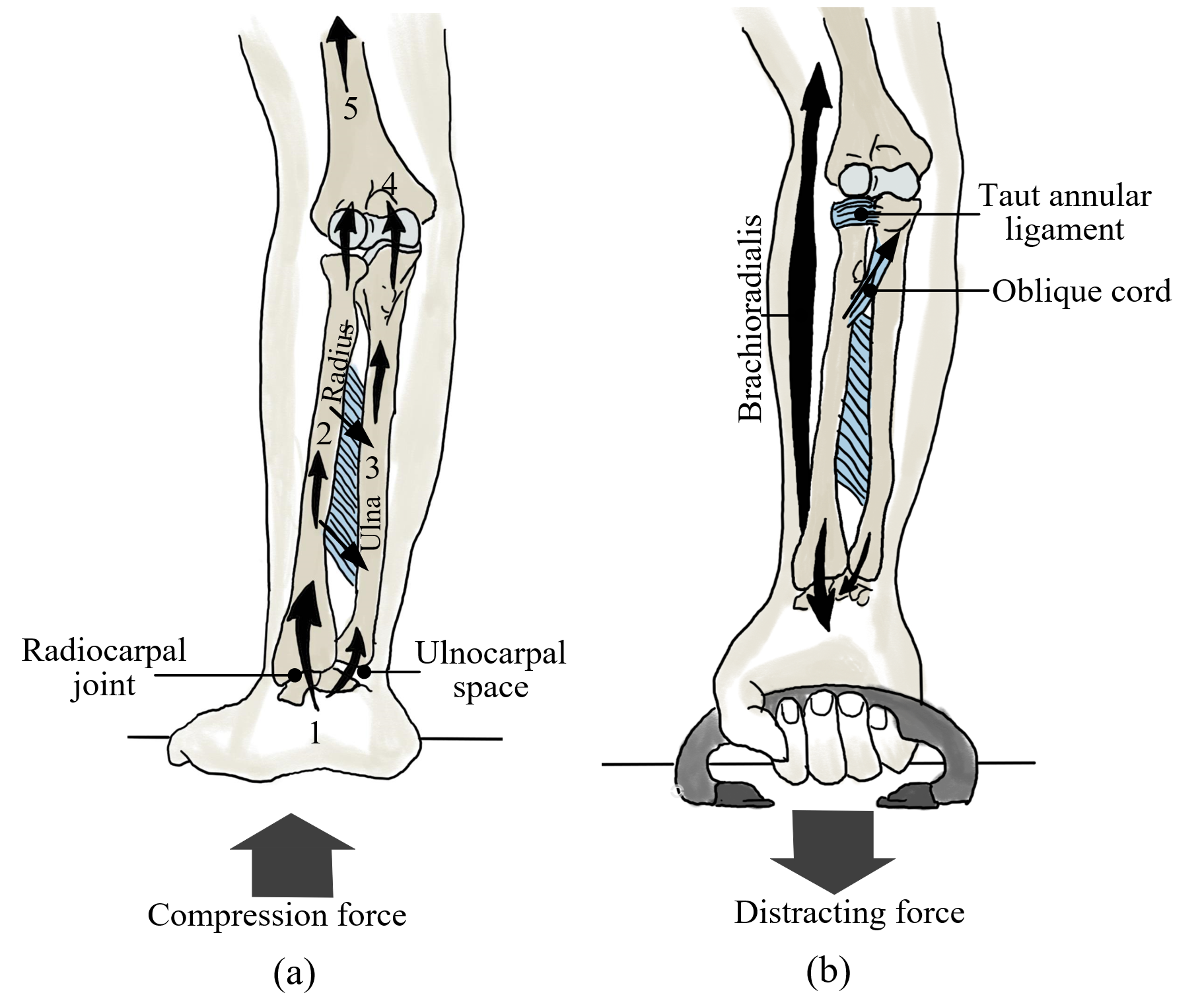}}
\caption{(a) A compression force applied on the hand is transmitted mainly through the wrist to the radius; (b) A distal-directed force is applied on the hand, predominantly through the radius.\cite{lockard2006clinical}}
\label{fig8.4}
\end{figure}

The interosseous membrane (IOM) plays a crucial role in connecting the ulna and radius throughout the length of the forearm (Fig.\ref{fig8.3}(a) (b))\cite{lockard2006clinical}. It is made up of three main parts: the distal membranous portion (DOB), the middle portion, and the proximal portion. The middle portion can be further divided into the central band (CB) and the accessory band (AB). The IOM performs several critical functions. First, it acts as a pivot for forearm rotation and connects the radius to the ulna. Second, it improves the stability of the DRUJ \cite{anderson2015role,werner2017role,hwang2019effects}, ensuring longitudinal stability for the forearm. Most importantly, research has indicated that the IOM can be viewed as a load transfer system that distributes the load from the radius to the ulna\cite{lockard2006clinical}. Approximately 80\% of the compression force crossing the wrist is directed through the radiocarpal joint (Fig.\ref{fig8.3}(a)), with the remaining 20\% crossing the distal side of the wrist via the soft tissues in the 'ulnocarpal space' \cite{burkett2022anatomy}. As shown in Fig.\ref{fig8.4}(a), the compression force acting on the radius from the wrist can be distributed to the ulna via the IOM, which helps reduce the load on the radial head and stabilizes the forearm against radioulnar bowing or splaying by drawing the ulna and radius towards the interosseous space. Similarly, As shown in Fig.\ref{fig8.4}(b), when a distracting force is applied to the distal radius from the wrist, this force tightens the fibres of the IOM, transferring the load to the ulna and limiting the load transferred to the proximal radius to be distributed across its limited articular surface area. As a result, IOM distributes the axial force from the radius to the ulna, and effectively disperses it across multiple joints (including the DRUJ, PRUJ, and humeroulnar joint), instead of transferring it directly to the humeroradial joint. This mechanism helps prevent dislocation or excessive stress in the humeroradial joint.

The DRUJ is a critical component of the forearm and wrist, and the triangular fibrocartilage complex (TFCC) plays a vital role in its stability (Fig.\ref{fig8.3}(c)) \cite{nakamura1996functional,szabo2006distal}. Composed of the palmar radioulnar ligament (PRUL), dorsal radioulnar ligament (DRUL), and extensor carpi ulnaris tendon (ECU), the TFCC helps maintain the proper alignment and function of the joint.

\begin{figure*}[htb]
\centerline{\includegraphics[width=1\textwidth]{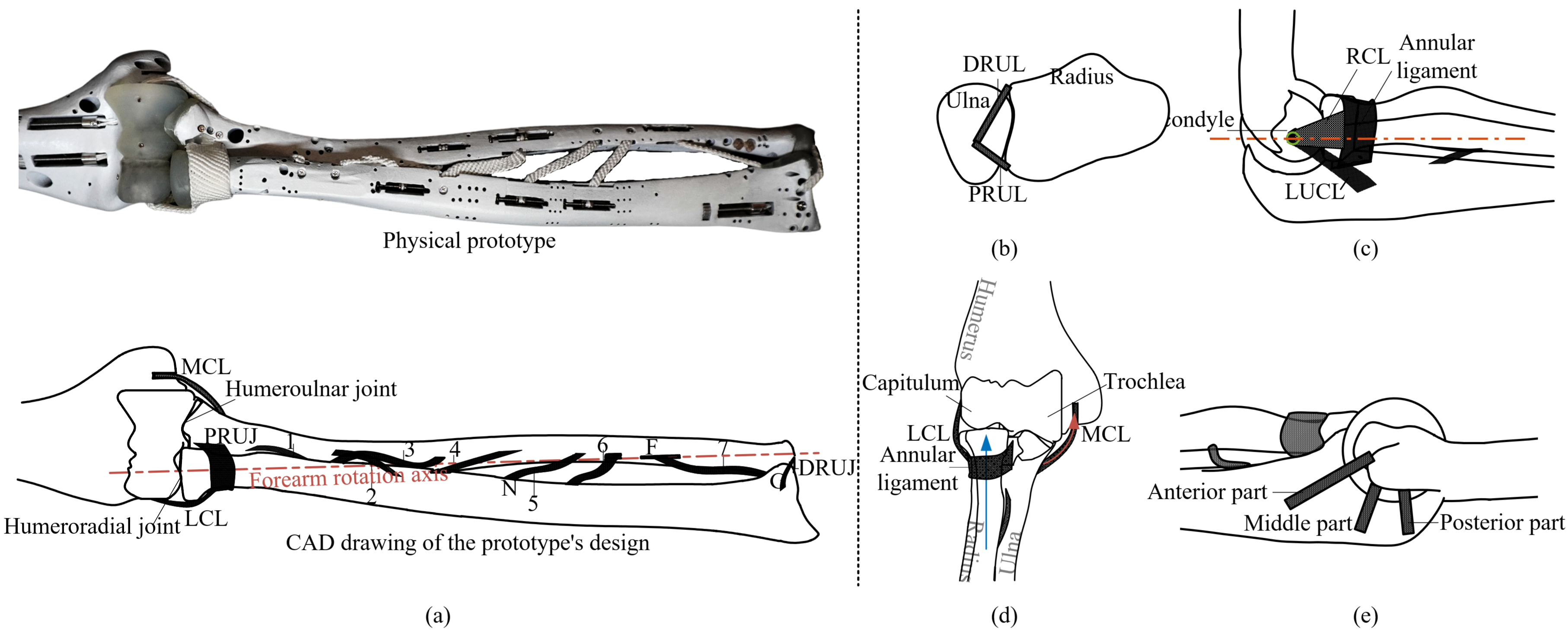}}
\caption{The design of the proposed robotic system (physical prototype and CAD drawing of the prototype's design). (a) Front view of the forearm; (b) TFCC structure; (c) Side view of the elbow, indicating the RCL and annular ligament; (d) Front view of the elbow, indicating the MCL, LCL and annular ligament; (e) Side view of the elbow, indicating the MCL.}
\label{fig5.0}
\end{figure*}

\subsection{Performance of the biological elbow and forearm}

The average percentage of total body weight and length of forearm is 1.72\% and 15.85\% \cite{plagenhoef1983anatomical}. Table \ref{tab2.4} presents the range of motion and output torques of the biological joints. Taking into account the dimensions and weight of the human arm, it becomes evident that the human arm can be regarded as an impressively powerful mechanism.

\begin{table}[htb]
\caption{Performance of biological elbow and forearm (data from \cite{dempster1955space,askew1987isometric}).}
\footnotesize
\begin{center}
\begin{tabular}{c c c}
\toprule
\makecell[c]{Motion group} & \makecell[c]{Range of motion} & \makecell[c]{Joint torque} \\
\midrule
Elbow Extension(-) / Flexion(+) & 0-142\degree & -41.3-71.1Nm\\
Forearm Pronation(-) / Supination(+) & -77\degree-113\degree & -7.16-8.93Nm\\
\bottomrule
\end{tabular}
\label{tab2.4}
\end{center}
\end{table}

This section concludes that contemporary robotic arm designs possess shortcomings, including the compromised safety of rigid robotic arms and instability in highly biomimetic variants. These issues are effectively resolved in the human arm, providing a blueprint for refining robotic arm design. Therefore, the forthcoming section will centre on replicating human arm characteristics to enhance robotic arm configuration.

\section{Biomimetic design of the Elbow-and-Forearm system}

The preceding section delineated the intricate structure and properties of the human arm. This section will introduce a novel, highly biomimetic robotic arm design, informed by the comprehensive understanding of bones, ligaments, and other soft tissues detailed earlier.

\begin{figure}[htp]
\centerline{\includegraphics[width=0.6\textwidth]{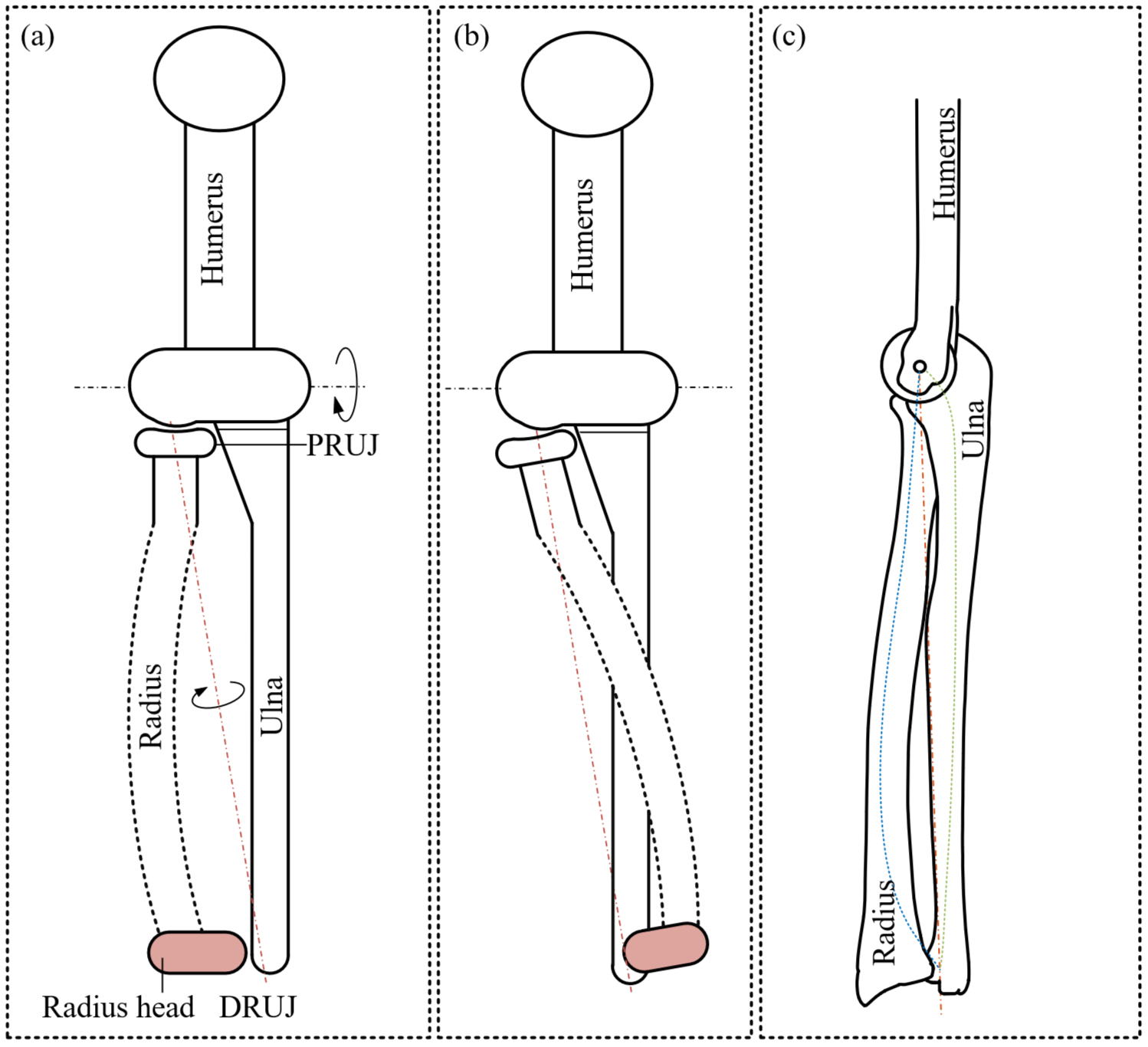}}
\caption{(a) Radius head allows an effective distance between radius and ulna; (b) The curved radius avoids interaction between the radius and ulna during forearm rotation; (c) The ulna is curved downwards near the elbow joint.}
\label{fig6.8}
\end{figure}

\subsection{Design of the skeletal structure}

In the proposed design, the elbow and forearm comprise the humerus, ulna, and radius, as shown in Fig. \ref{fig5.0}(a). Each joint within the skeletal structure is characterized by a thin layer of cartilage coating the contact surface. Additionally, the ligament systems encompassing TFCC, IOM, LCL, MCL, and the annular ligament are replicated within the robotic elbow and forearm.

The primary motion of the ulna is rotation around the humerus, which can be simplified to a hinge joint. The humeroulnar joint can achieve sufficient lateral and axial stability by relying on the MCL, LCL, and olecranon process. The radius can rotate relative to the humerus and rotate around the fixed ulna around the axis (shown in red in Fig. \ref{fig6.8}(b)) to achieve forearm rotation. Their unique geometry enables a wide range of motion in forearm rotation. The distal radial head, shown in red in Fig. \ref{fig6.8}(a), maintains an effective distance between the radius and ulna, preventing interference and maximizing the range of motion (Fig. \ref{fig6.8}(b)). This increased distance also enhances output torque during forearm rotation. The curved middle portion of the radius (Fig. \ref{fig6.8}(b)) and the downward curve of the ulna near the elbow joint (Fig. \ref{fig6.8}(c)) create space between them and the rotation axis (dashed red line), allowing the radius to rotate around the ulna without contact interference. This configuration significantly enhances the mobility of the radius. However, this increased mobility also causes instability in the radius across various directions. Consequently, in our design, we will incorporate essential soft tissues, drawing from anatomical features, to achieve stability for the forearm.

\subsection{Design of the soft tissues}


The design of soft tissues was optimised, congruent to human anatomical structures, facilitating their emulation through engineered materials. Fig. \ref{fig5.0} demonstrates the spatial distribution and the architectural design of these soft tissues.

\subsubsection{MCL}

To mimic the hinge function of a biological elbow in the robotic counterpart, the MCL complex is subdivided into three segments: anterior, middle, and posterior, as shown in Fig. \ref{fig5.0}(e). The anterior segment originates above the elbow rotation centre, the middle segment at the centre, and the posterior segment below it. This arrangement allows the middle segment to offer stability throughout the elbow rotation while the tension in the anterior and posterior segments increases significantly near full extension and flexion, limiting the maximum motion range. Video 1.4 in the supplementary material presents the MCL during elbow flexion/extension. By replicating the biological MCL complex, the robotic elbow attains joint stability and a range of motion comparable to that of a human elbow joint.

\subsubsection{Annular ligament}

The annular ligament is essential for stabilizing the PRUJ in the robotic forearm. As shown in Fig.\ref{fig5.0}(e) and (d), it comprises multiple fibres woven into a short circular tube, originating from the ulna, encircling the radial head, and reinserting into the ulna.

\subsubsection{LCL}

In the robotic elbow, the LCL comprises the RCL and LUCL (Fig. \ref{fig5.0}(c)). The RCL connects the lateral epicondyle to the annular ligament, while the LUCL links the lateral epicondyle to the ulna. Together with the MCL, they hinge the forearm to the humerus, contributing to the elbow joint's stability, as depicted in Fig. \ref{fig5.0}(d).

\subsubsection{TFCC}

The TFCC in the robotic forearm (Fig. \ref{fig5.0}(b), consisting of DRUL and PRUL, originates from the ulna and inserts into the radius. It stabilizes the DRUJ, while the annular ligament secures the PRUJ, enabling the radius to be hinged to the ulna.

\subsubsection{IOM}

\begin{figure}[htb]
\centerline{\includegraphics[width=0.6\textwidth]{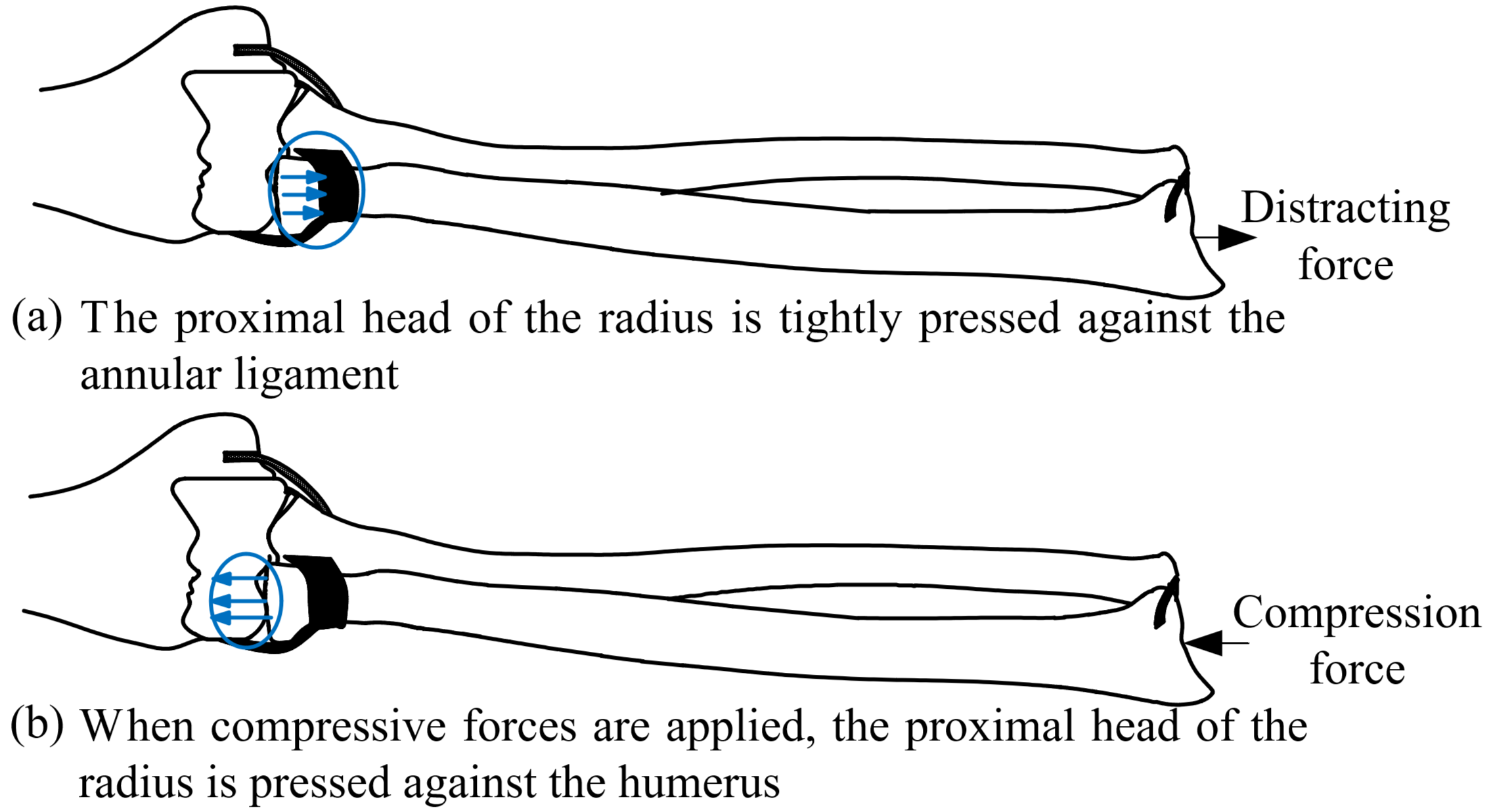}}
\caption{The situation prior to the installation of IOM. (a) Distracting forces are applied to the radius distal head, and the proximal head of the radius is tightly pressed against the annular ligament. (b) Compressive forces are applied, and the proximal head of the radius is pressed against the humerus}
\label{fig5.21}
\end{figure}

Fig.\ref{fig5.0}(a) illustrates the arrangement of the seven major portions of the IOM in the proposed design. The IOM can reduce friction in the humeroradial joint and between the annular ligament and radial head, decreasing resistance during forearm rotation. As shown in Fig.\ref{fig5.21}, without the IOM, the annular ligament and DRUL/PRUL restrict the radius's axial movement, generating friction when distracting forces are applied to the radius distal head. The radius proximal head is pressed against the humerus, resulting in significant friction in the humeroradial joint when compression forces are applied. The IOM can distribute these forces across its seven portions, reducing friction and enabling smoother forearm rotation.

This section delves into the application of engineering materials, mirroring human arm constituents such as bones, ligaments, and cartilage, in the design of the robotic arm. Preliminary tests indicate that the arm can replicate the motion functions of the human counterpart and maintain joint stability. The ensuing section will decode the biological principles inherent in these designs.

\section{Modeling and Stability Analysis of the Radius-Ulna Joints}

In the previous section, the robotic forearm and elbow, inspired by the human skeletal ligament system, were introduced. In the design, the ulna is firmly hinged to the humerus due to the MCL and LCL, providing considerable stability. The radius has two degrees of freedom, enabling a wide range of motion, which makes it more susceptible to dislocation compared to the ulna. The key to stabilizing the radius as it rotates around the ulna is to hinge it on its rotation axis, connecting it to the stable ulna. As shown in Fig. \ref{fig5.0}(a), the forearm rotation axis (red dashed line) passes through the rotation centres of the humeroradial joint, PRUJ, and DRUJ. The stability of these joints is achieved through mechanisms formed by soft tissues and joint surfaces. {Several mechanical features and principles, derived from studying the human arm, have been identified as potentially contributing to the high stability of the radius.} These include the ball and socket structure of the humeroradial joint, TFCC stabilising the DRUJ, improving forearm stability through IOM, and variation in MCL strain during elbow movement. This section will theoretically analyze how the proposed design's mechanisms anchor the radius to the axis of forearm rotation and sustain stability.
 
\subsection{Ball and socket structure of the humeroradial joint}

The humeroulnar joint and PRUJ work in conjunction to stabilize the proximal radius. The interplay of the annular ligament, radial head, capitulum, and RCL aids in maintaining the radial and axial position of the proximal radius. The PRUJ's rotation centre, situated on the forearm rotation axis, is constrained by the annular ligament and RCL (Fig. \ref{fig5.0}(a)), assisting in the prevention of lateral dislocation of the radius. The humeroradial joint, located between the radial head and capitulum, operates as a ball-and-socket joint, with its rotation centre also residing on the forearm rotation axis. As shown in Fig. \ref{fig5.0}(c) and (d), the RCL and annular ligament apply pressure to the socket (radial head), which in turn pushes it against the ball (capitulum), thus enhancing the lateral stability of the radius.

\begin{figure}[htb]
\centerline{\includegraphics[width=0.5\textwidth]{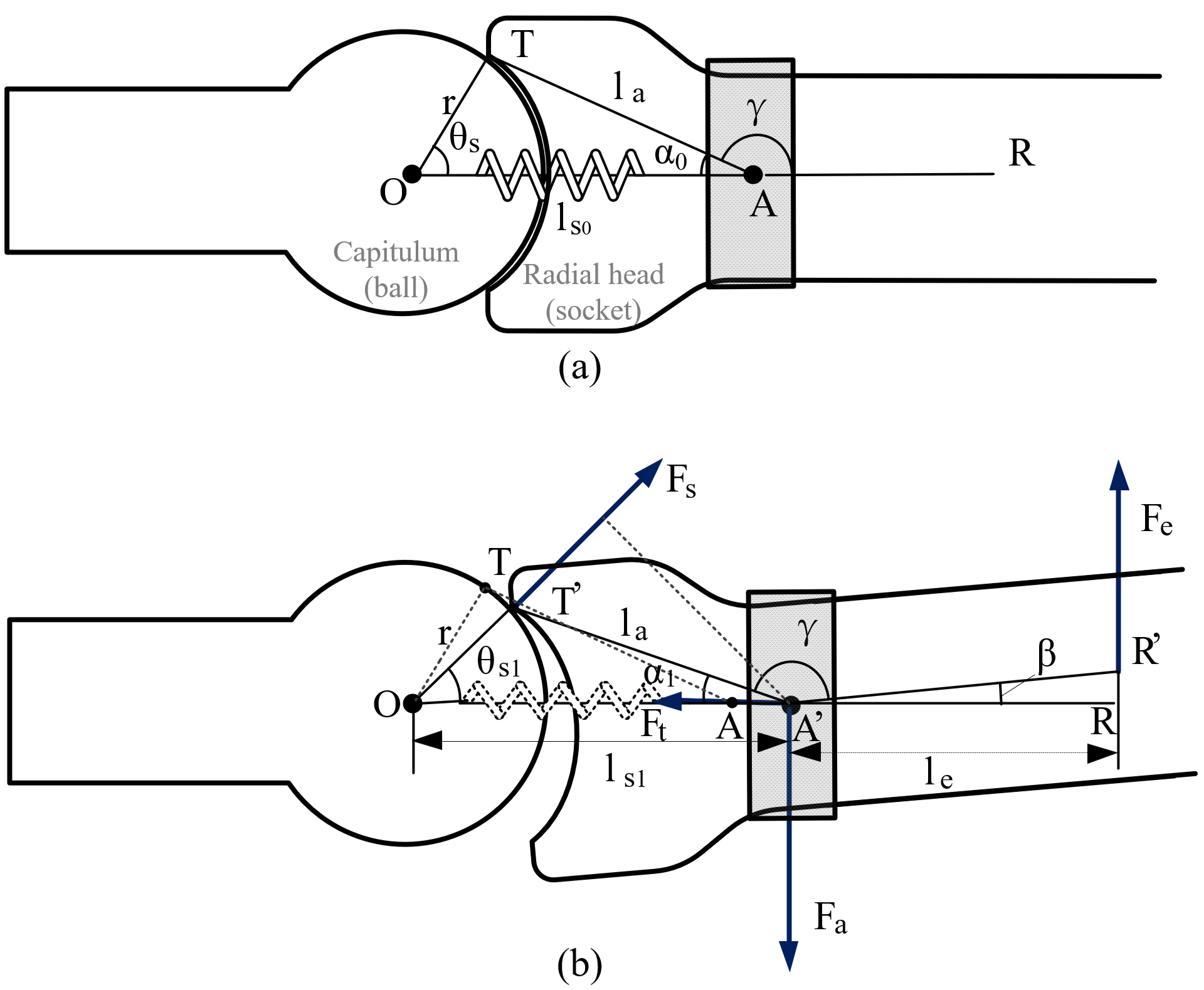}}
\caption{The simplified diagram of the force on the humeroradial joint. (a) Initial stage; (b) Stage when the joint is dislocated.}
\label{fig6.10a}
\end{figure}

The humeroradial joint can be simplified as Fig. \ref{fig6.10a}(a). Point $A$ is the articulation point of the annular ligament and radius. Point $O$ is the spherical centre of the capitulum. The LCL can be simplified to a spring with high stiffness, presented by $OA$. Point $T$ is the contacted endpoint of the radial head and capitulum. Since the radial head is not a complete socket, there is an initial angle between $TO$ and the horizontal line, denoted as $\theta_s$. The joint is more stable as $\theta_{s}$ increases, but the range of motion will be limited and vice versa.

When an external force $F_{e}$ is applied to the distal end of the radius, only lateral forces are considered, as shown in Fig. \ref{fig6.10a}(b), the humeroradial joint will start to dislocate. The joint contact point slides from $T$ to $T'$. As the radial head is retained by the annular ligament, the radius and the annular ligament can be approximated as hinged at point $A$. The annular ligament is fixed to the ulna, and assuming that the ulna is fixed, the annular ligament can only move a small distance in the horizontal direction. During the dislocating of the humeroradial joint, the position of $TA$ will move to $T'A'$. The LCL will be stretched to $OA'$. Radius will deflect. The relationship between $F_{e}$ and the elongation of LCL $\Delta l_{s}$ will be calculated.

When $F_{e}$ is applied to the radius, as shown in Fig. \ref{fig6.10a}(b), the annular ligament will provide support force $F_{a}$. There will be a support force $F_{s}$ from the contact point $T'$, and the tensile force from the LCL through the annular ligament  $F_{t}$. According to the force balance:
\begin{equation}
\begin{cases}
F_{s} cos{\theta_{s1}}=k \Delta l_{s} \\
F_{s} sin{\theta_{s1}}+F_{e}=F_{a} \\
F_{s} l_{s1} sin{\theta_{s1}}=F_{e} l_{e} \\
\end{cases}
\label{eq6.4}
\end{equation}

Where $\theta_{s1}$ is the angle between $TO$ and the horizontal, $l_{s1}$ is the length of $OA'$, $k$ is the elasticity coefficient of the ligament, $l_{e}$ is the moment arm of the external force, $\Delta l_{s}$ can be calculated as:

\begin{equation}
\Delta l_{s}=l_{s1}-l_{s0}
\label{eq6.7}
\end{equation}

Where $l_{s0}$ is the initial length of $OA$.

Combine equations \ref{eq6.4} and \ref{eq6.7}, the relation between $F_{e}$, $\Delta l_{s}$ and $\theta_{s1}$ can be obtained as function:

\begin{equation}
F_{e}=f_{1}(\Delta l_{s}, \theta_{s1})
\label{eq6.8}
\end{equation}

According to Fig. \ref{fig6.10a}(b), in $\Delta T'OA'$, there are:
\begin{equation}
\begin{cases}
\alpha_{1}=\pi-\gamma-\beta\\
cos\alpha_{1}=\frac{l_{s1}^{2}+l_{a}^{2}-r^{2}}{2l_{s1}l_{a}}\\
\frac{r}{sin\alpha_1}=\frac{l_a}{sin\theta_{s1}}
\end{cases}
\label{eq6.9}
\end{equation}

Where $\alpha_1$ is the angle between $TA$ and the horizontal line changes from $\alpha_0$. $\gamma$ is the angle between $TA$ and the radius axis $AR$, which is constant. $\beta$ is the radius deflect angle. $l_a$ is the length of $\bar{TA}$ and $\bar{T'A'}$. $r$ is the length of $OT$.

According to equations \ref{eq6.7} and \ref{eq6.9}, the relation between $\beta$ and $\Delta l_{s}$ can be obtained as function:
\begin{equation}
\beta=f_2(\Delta l_{s})
\label{eq6.12}
\end{equation}

According to equation \ref{eq6.9}, the relation between $\theta_{s1}$ and $\beta$ can be obtained as a function:
\begin{equation}
\theta_{s1}=f_{3}(\beta)
\label{eq6.13}
\end{equation}

\begin{figure}[htb]
\centerline{\includegraphics[width=0.55\textwidth]{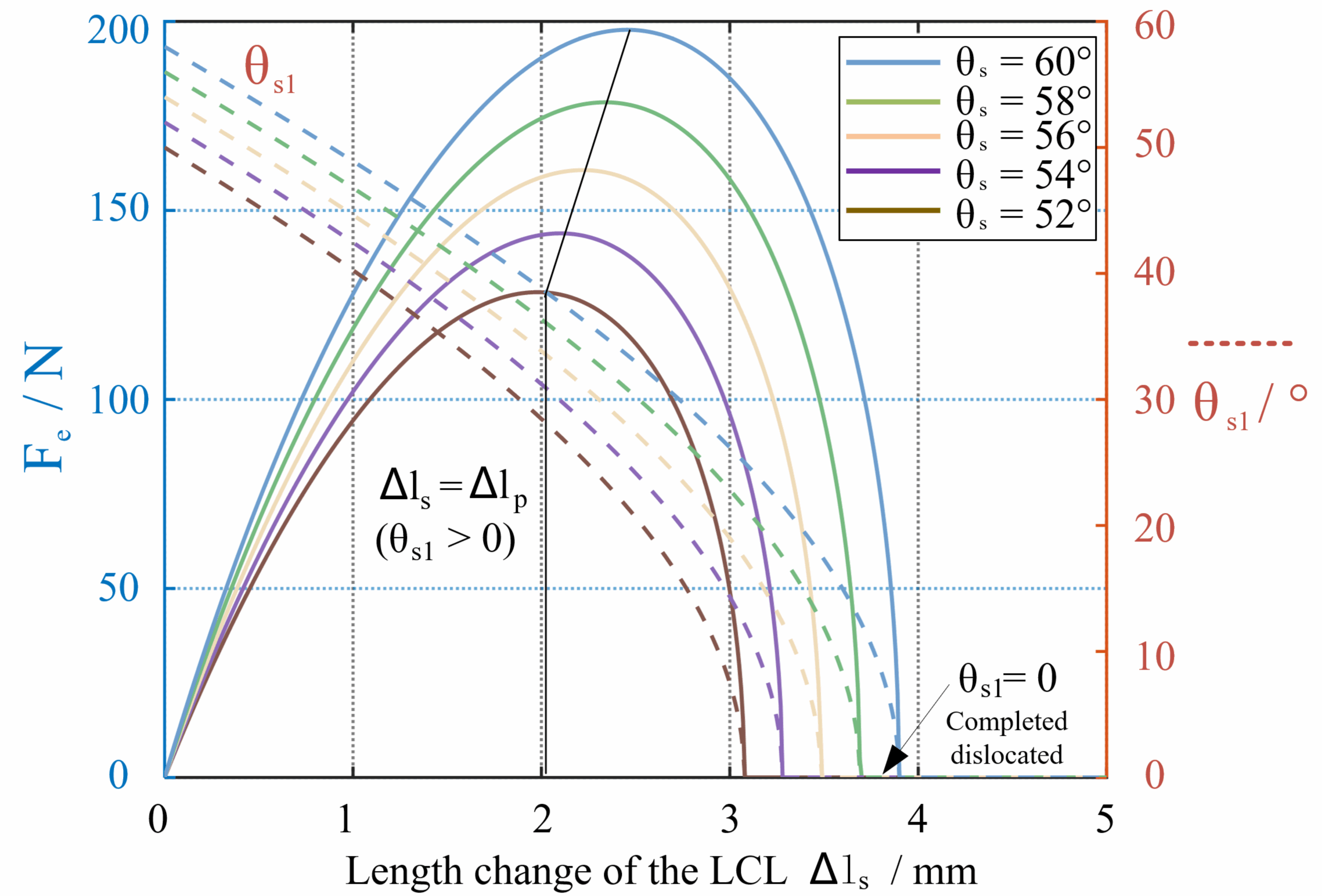}}
\caption{The simulation result of the relationship between the force applied on the distal radius and the length change of LCL.}
\label{fig6.10b}
\end{figure}

According to equations \ref{eq6.8}, \ref{eq6.12} and \ref{eq6.13}, the relation between $F_{e}$ and $\Delta l_{s}$ can be obtained. The similation results between $F_{e}$ and $\Delta l_{s}$, $\theta_{s1}$ and $\Delta l_{s}$ with different $\theta_s$ are shown in Fig. \ref{fig6.10b}. The solid curves in the figure show that increasing the angle $\theta_s$ between the horizontal line and $TO$ enhances the joint's capacity to withstand external forces $F_{e}$.

When the joint is dislocated by $F_{e}$, as shown in the solid curves in Fig. \ref{fig6.10b}, the force $F_{e}$ required will increase first and then decrease after reaching the peak value when the LCL is stretched as $\Delta l_{s}$ increases. When $\Delta l_{s}<\Delta l_{p}$ ($\Delta l_{p}$ denotes the value of $\Delta l_{s}$ when $F_{e}$ reach the peak value), $F_{e}$ needs to be increased continuously to make the joint dislocation more severe. At this stage,  $\theta_{s1}$ \textgreater 0, as shown in the dashed line in Fig. \ref{fig6.10b}, the joint may recover automatically if the external force is withdrawn. When the LCL ligament stretches to $\Delta l_{s}>\Delta l _{p}$, even if $F_{e}$ decreases or is removed, the joint dislocation will deteriorate, the joint may dislocate automatically until $\theta_{s1}=0$. So the joint dislocation happens when $\Delta l_{s}=\Delta l_{p}$, even if $\theta_{s1}>0$ in this stage, the joint is dislocated.

\subsection{TFCC stabilize the DRUJ}

The TFCC structure (Fig. \ref{fig5.0}(b)) constrains the DRUJ's rotation centre and, in conjunction with the PRUJ, enables the radius to maintain initial stability. Notably, the rotation centre of the TFCC and the joint surface rotation centre (on the forearm rotation axis) are not aligned. To address this misalignment, the ECU and PCU tendons, which actuate the hand, prompt the DRUL or PRUL to encircle them as the radius rotates. Consequently, even though the TFCC's rotation centre does not align with the joint rotation centre, the TFCC can still sustain tension and restrict the joint rotation centre. 

\begin{figure}[htb]
\centerline{\includegraphics[width=0.68\textwidth]{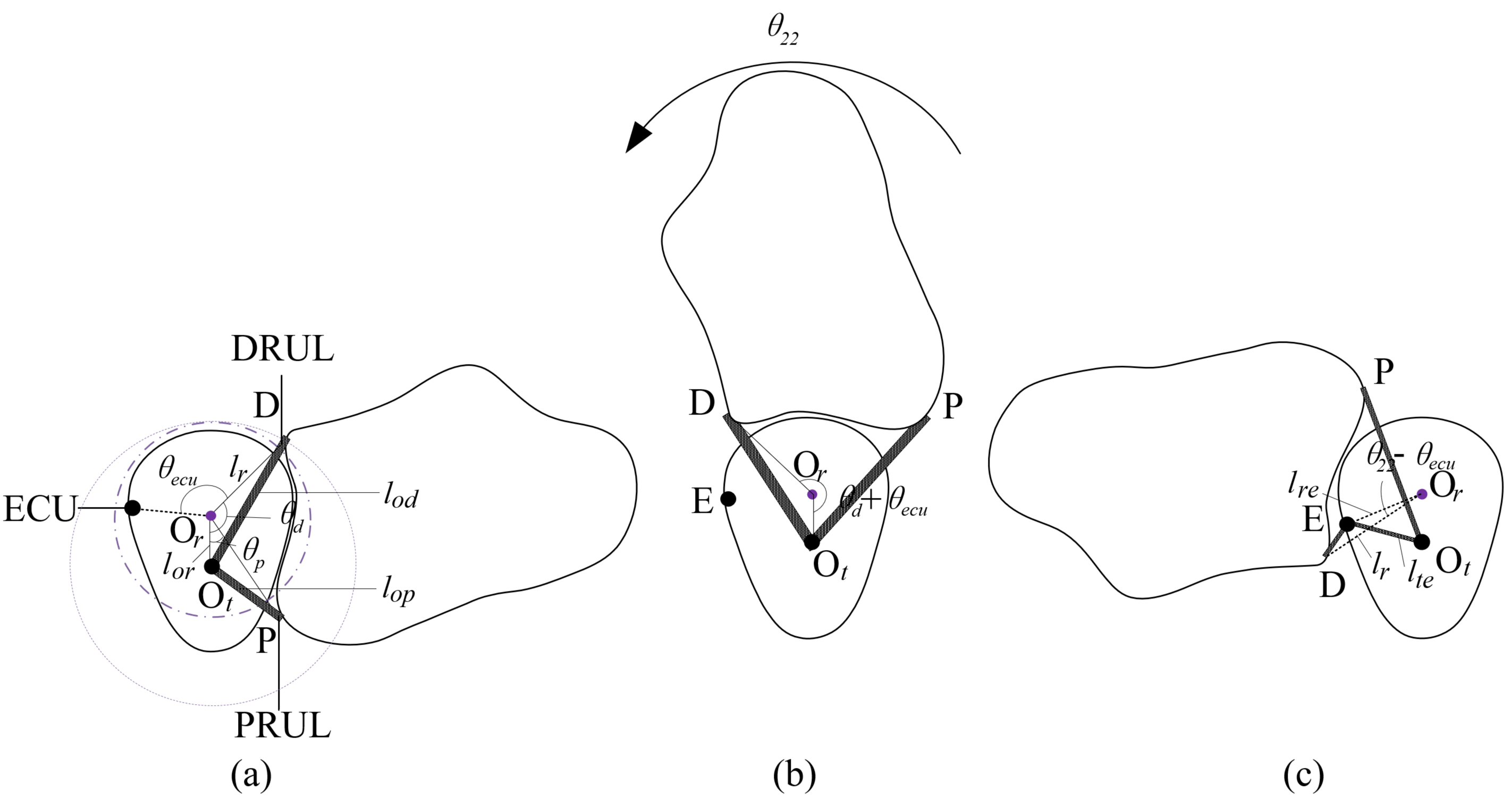}}
\caption{The simplified diagram of the TFCC structure when the forearm (a) fully supinated, (b) during pronating, and (c) fully pronated.}
\label{fig6.4}
\end{figure}

The simplified diagram of this structure during forearm rotation is shown in Fig. \ref{fig6.4}. DRUL is in contact with the ECU when the forearm rotates at $\theta_{22}=\theta_{ecu}$($\theta_{22}$ is the joint position). Point $D$ and point $P$ are the joint contact edge points. Point $E$ represents the location of ECU. $O_t$ is the rotation centre of TFCC. $O_r$ is the rotation centre of the DRUJ joint contact surface.

Before DRUL contacts with the ECU ($\theta_{22}<\theta_{ecu}$), the relationship between the length changes of DRUL $\Delta_{d}$, PRUL $\Delta_{p}$,  and $\theta_{22}$ can be calculated as:
\begin{equation}
\begin{cases}
\Delta_{d}=\sqrt{(l_{r}^{2}+l_{or}^{2}-2l_{r} l_{or} cos(\theta_{d}+\theta_{22}))}-l_{od}\\
\Delta_{p}=\sqrt{(l_{r}^{2}+l_{or}^{2}-2l_{r} l_{or} cos(\theta_{p}+\theta_{22}))}-l_{op}
\end{cases}
\label{eq6.1}
\end{equation}

Where $l_r$ is the length of the $O_rD$ and is a constant. $l_{or}$ is the length of $O_t O_r$. $\theta_{d}$ is the angle of $\angle DO_rO_t$, it will increase to $\theta_{d}+\theta_{22}$ when the joint rotates. $l_{od}$ is the length of $O_tD$ when the forearm is in its initial position, i.e. the initial length of DRUL. $l_{op}$ is the length of $O_tP$, which is the initial length of PRUL. $\theta_{p}$ is the angle of $\angle PO_rO_t$, it will increase to $\theta_{p}+\theta_{22}$ when the joint rotates.

After DRUL contacts with the ECU ($\theta_{22}\geq\theta_{ecu}$), $\Delta_{d}$ varies with $\theta_{22}$ as:
\begin{equation}
\Delta_{d}=\sqrt{(l_{r}^{2}+l_{re}^{2}-2l_{r} l_{re} cos(\theta_{22}-\theta_{ecu}))}+l_{te}-l_{od}
\label{eq6.3}
\end{equation}

Where $l_{re}$ is the length of $O_rE$, $l_{te}$ is the length of $O_tE$ (Fig. \ref{fig6.4}(c)).

\begin{figure}[htb]
\centerline{\includegraphics[width=0.45\textwidth]{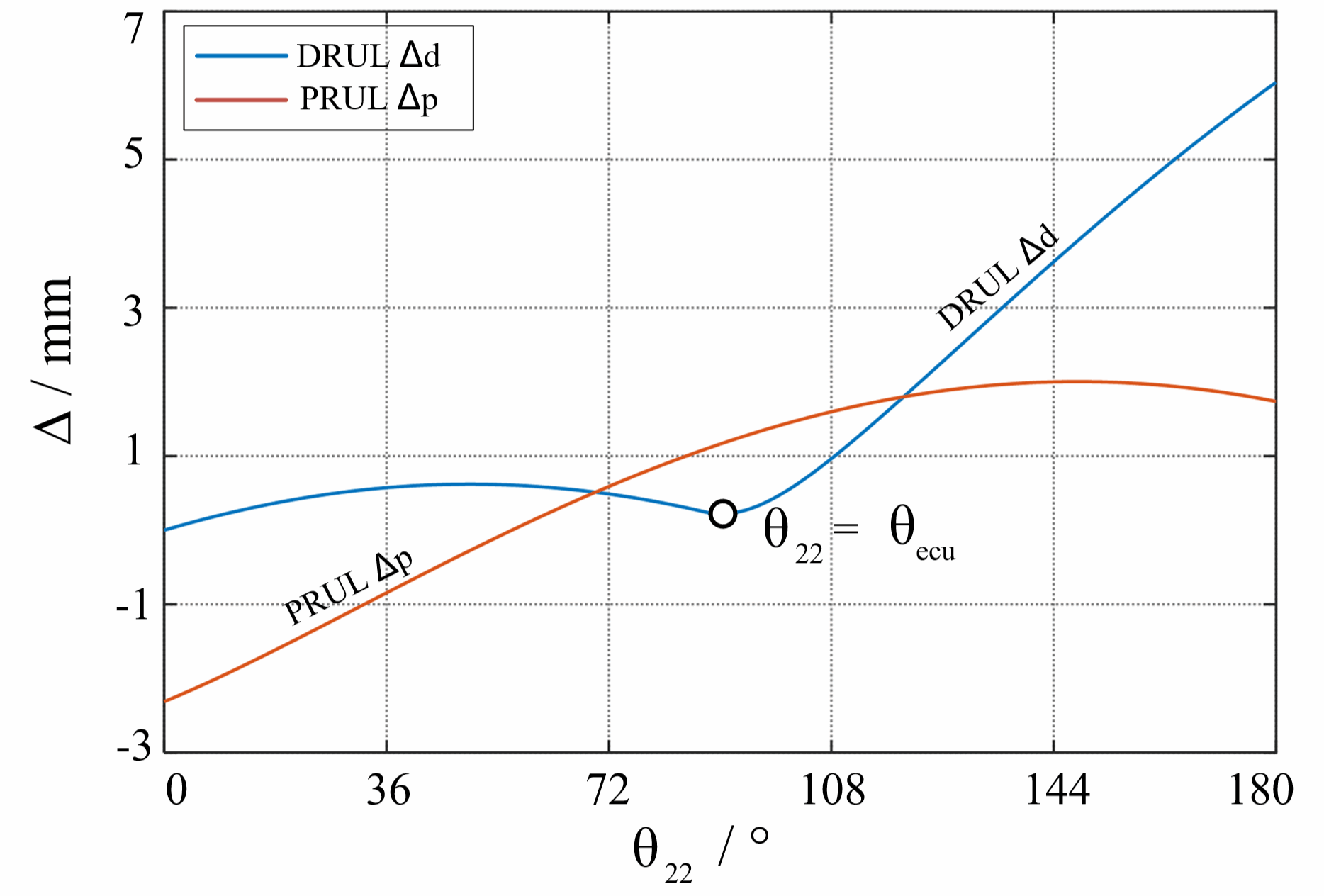}}
\caption{The simulation result of relation between $\Delta_{d}$, $\Delta_{p}$ and $\theta_{22}$.}
\label{fig6.5}
\end{figure}

In the design, the length of the DRUL is adjusted to ensure tension when it contacts with the ECU. The relationship between the changes in the lengths of the DRUL and PRUL and the joint angle ($\theta_{22}$) is illustrated in Fig. \ref{fig6.5}. It can be observed that when $\theta_{22}<\theta_{ecu}$, the DRUL (blue) is almost not stretched. When it comes into contact with the ECU, it rapidly stretches, effectively limiting the maximum position of $\theta_{22}$. During forearm rotation, the PRUL (red) is initially relaxed and then stretched, with the total amount of relaxation and stretch not exceeding 2 mm. Thus, {the TFCC structure is not able to stabilise the DRUJ} and other soft tissues, such as the IOM, further measures are required to enhance stability.

\subsection{Improving forearm stability through IOM}\label{section7.2}

\begin{figure*}[htb]
\centerline{\includegraphics[width=1\textwidth]{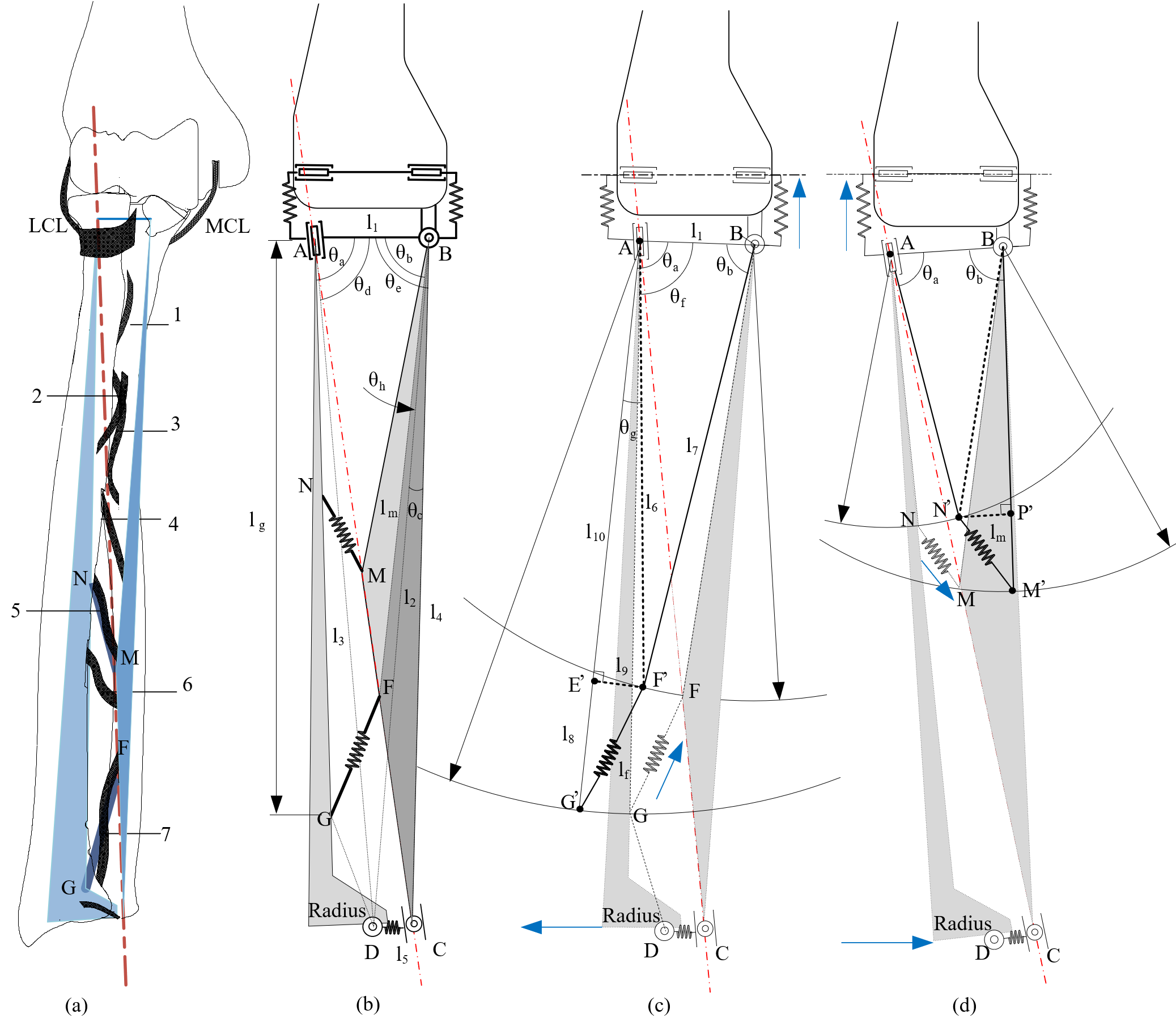}}
\caption{(a) ligaments in IOM; (b) The simplified structure of ligaments 5 and 7; (c) The simplified diagram when the radius is under lateral external force to the left; (d) The simplified diagram when the radius is under lateral external force to the right.}
\label{fig6.12}
\end{figure*}

While the annular ligament, LCL, and TFCC structures offer initial stability to the radius, the TFCC does not maintain constant tension during forearm rotation, suggesting limited stability in the DRUJ and PRUJ. Besides these structures, the IOM significantly contributes to forearm stabilization by serving as a hinge between the radius and ulna. The membrane features distinct bundles with diverse orientations, enhancing axial and lateral stability. Since the membrane bundles' insertion points on the ulna and radius reside on the forearm rotation axis (Fig. \ref{fig8.3}(b)), the membrane does not generate resistance during forearm rotation. This enables a broad range of motion without sacrificing stability.

Fig. \ref{fig6.12}(a) depicts the forearm with intact MCL, LCL, TFCC, IOM, and the annular ligament. When a lateral force is exerted on the distal end of the radius, the IOM aids in counteracting the external force and transfers it to the LCL and MCL. This subsection will explore the mechanism by which the IOM assists in resisting lateral external forces.

Under external lateral forces, the IOM bundles in the same inclined direction transfer force through a similar mechanism. To examine the stability offered by the IOM from various directions, two IOM bundles in distinct orientations were chosen for analysis, specifically ligament 5 and ligament 7, as displayed in Fig. \ref{fig6.12}(a). The derivation process can be directly applied to other IOM ligaments.

With only ligament 5 and ligament 7 retained, the forearm can be simplified to the configuration depicted in Fig. \ref{fig6.12}(b). The forearm rotation axis (red line) passes through the insertion points of ligament 5 and ligament 7 on the ulna, as illustrated in Fig. \ref{fig6.12}(a). The simplified representation in Fig. \ref{fig6.12}(b) displays the characteristics and parameters defining the ligaments' positions, while the remaining structures are simplified. The strain in ligaments 5 and 7 stays constant during forearm rotation, ensuring the simplified diagram accurately represents the geometric relations even as the forearm rotates. This allows the IOM to stabilize the forearm by transmitting lateral forces.

As illustrated in Fig. \ref{fig6.12}(b), quadrilateral $ABCD$ can be simplified into a planar configuration with hinges $A$, $B$, $C$, and $D$ free to rotate. The radius and interosseous ligaments 5 and 7 are permitted to rotate around the axis $AC$. Segment $CD$ represents the TFCC structure with a constant length. Consequently, $ABCD$ can be regarded as an unstable quadrilateral with fixed side lengths. When an external force is applied to the distal radius, the ulna undergoes rotational movement as the radius rotates. First, the angular relationship between the ulna's ($BC$) rotation and the radius's ($AD$) rotation in the plane will be calculated.

In $\Delta ABD$, according to the cosine and sine law, there has:
\begin{equation}
\begin{cases}
l_{2}^{2}=l_{1}^{2}+l_{3}^{2}-2l_{1} l_{3}cos\theta_{d} \\
\frac{l_{2}}{sin\theta_{d}}=\frac{l_{3}}{sin\theta_{h}}
\end{cases}
\label{eq6.14}
\end{equation}

Where $l_{1}$, $l_{2}$, and $l_{3}$ represent the lengths of segments $AB$, $BD$, and $AD$, respectively. Both $l_{1}$ and $l_{3}$ remain constant. $\theta_d$ represents $\angle BAD$, which is variable. $\theta_h$ represents $\angle ABD$.

So, $l_{2}$ and $\theta_{h}$ can be obtained. In $\Delta BCD$, according to cosine low:
\begin{equation}
l_5^2=l_2^2+l_4^2-2l_2 l_4cos\theta_c
\label{eq6.16}
\end{equation}

Where, $l_4$ and $l_5$ represents segments $BC$ and $CD$, both are constant. $\theta_c$ represents $\angle CBD$.

$\theta_c$ can be obtained. $\theta_e$ ($\angle ABC$) can be calculated as:
\begin{equation}
\theta_e=\theta_h+\theta_c
\label{eq6.17}
\end{equation}

Combine equations \ref{eq6.14} to \ref{eq6.17}, the relation between $\theta_e$ and $\theta_d$ can be obtained, i.e. the angular relationship between ulna's consequent rotation when the radius is rotated. It can be denoted as a function:
\begin{equation}
\theta_e=f_{RU}(\theta_d)
\label{eq6.18}
\end{equation}

In Fig. \ref{fig6.12}(b), ligament 7 is denoted by $FG$, while ligament 5 is represented by $MN$. The insertion points of ligaments 7 and 5 on the ulna are labeled as $F$ and $M$, respectively, and their respective insertion points on the radius are designated as $G$ and $N$. Both $\Delta BFC$ and $\Delta BMC$ remain undeformed during ulna deflection, and the angles $\angle CBF$ and $\angle CBM$ are constant. Similarly, $\Delta ADG$ and $\Delta ADN$ do not deform as the radius deflects, maintaining constant angles $\angle DAG$ and $\angle DAN$. The angles $\theta_a$ (corresponding to $\angle BAG$ for ligament 7 or $\angle BAN$ for ligament 5; in Fig. \ref{fig6.12}(b), it represents $\angle BAG$) and $\theta_b$ (referring to $\angle ABF$ for ligament 7 or $\angle ABM$ for ligament 5; in Fig. \ref{fig6.12}(b), it represents $\angle ABF$) can be calculated as follows:
\begin{equation}
\begin{cases}
\theta_a=\theta_d+\angle DAG(N) \\
\theta_b=\theta_e-\angle CBF(M)
\end{cases}
\label{eq6.19}
\end{equation}

Combined with equation \ref{eq6.18}, the relation between $\theta_a$ and $\theta_b$ can be obtained as a function:
\begin{equation}
\theta_b=f_{ab}(\theta_a)
\label{eq6.21}
\end{equation}

When a lateral external force is applied to the distal end of the radius in a leftward direction (typically originating from the hand), the radius deflects clockwise around point $A$, as illustrated in Fig. \ref{fig6.12}(c). Due to the TFCC structure ($CD$), the ulna also experiences deflection around point $B$. However, the MCL becomes reinforced and resists the ulna's deflection. Given the high strength and stiffness of the MCL, it mitigates the ulna's deflection, thereby maintaining the forearm's stability.

During the clockwise deflection of the radius and ulna, the quadrilateral $ABCD$ undergoes deformation. As illustrated in Fig. \ref{fig6.12}(c), the quadrilateral $ABFG$ also experiences deformation, transforming from the dashed line to the solid line $ABF'G'$. Based on equation \ref{eq6.21}, the relationship between $\theta_a$ and $\theta_b$ is established. Consequently, ligament $FG$ is stretched to $F'G'$. In $\Delta AF'B$, according to cosine and sine law, it has:
\begin{equation}
\begin{cases}
l_6^2=l_1^2+l_7^2-2l_1 l_7cos\theta_b \\
\frac{l_6}{sin\theta_b}=\frac{l_7}{sin\theta_f}
\end{cases}
\label{eq6.22}
\end{equation}

Where, $l_6$ represents the length of $AF'$, while $l_7$ denotes the constant length of $BF'$. $\theta_f$ corresponds to the angle $\angle BAF'$. 

$l_6$ and $\theta_f$ can be obtained.

In the right triangle $\Delta AE'F'$, it has:
\begin{equation}
\begin{cases}
\theta_g=\theta_a-\theta_f \\
l_9=l_6 sin\theta_g \\
l_{10}=l_6 cos\theta_g
\end{cases}
\label{eq6.24}
\end{equation}

Where, $l_9$ and $l_{10}$ corresponds to the length of $E'F'$ and $AE'$, respectively. $\theta_{g}$ represents the angle $\angle F'AE'$.

In a right-angled triangle $\Delta E'F'G$, it has:
\begin{equation}
\begin{cases}
l_8=l_g-l_{10}\\
l_f=\sqrt{l_9^2+l_8^2}
\end{cases}
\label{eq6.27}
\end{equation}

Where, $l_8$ represents the length of $E'G'$, while $l_f$ corresponds to the length of $F'G'$. $l_g$ denotes the length of $AG'$.

Combining equations \ref{eq6.21} to \ref{eq6.27}, the relationship between $l_f$ and $\theta_a$ can be derived, denoted as $l_f = f_{fg}(\theta_a)$. This expression represents the connection between the length of ligament 7 and the deflection angle of the radius.

As depicted in Fig. \ref{fig6.12}(d), when a lateral external force is applied to the distal end of the radius in the rightward direction, the LCL ligament is strengthened and counters the counterclockwise deflection of the radius. This deflection also causes the ulna to deflect via the PRUJ. Quadrilateral $ABCD$ is deformed, transforming quadrilateral $ABMN$ from the dashed line state to the solid line state, represented as $ABM'N'$. According to equation \ref{eq6.21}, the relationship between $\theta_a$ and $\theta_b$ is established, and ligament $MN$ is stretched to $M'N'$, increasing its strain. This strain resists deflection of the radius and ulna, contributing to the overall stability. Using a similar methodology, the relationship between $l_m$ (length of $M'N'$) and $\theta_a$ can be derived as function $l_m = f_{mg}(\theta_a)$.

\begin{figure}[htb]
\centerline{\includegraphics[width=0.55\textwidth]{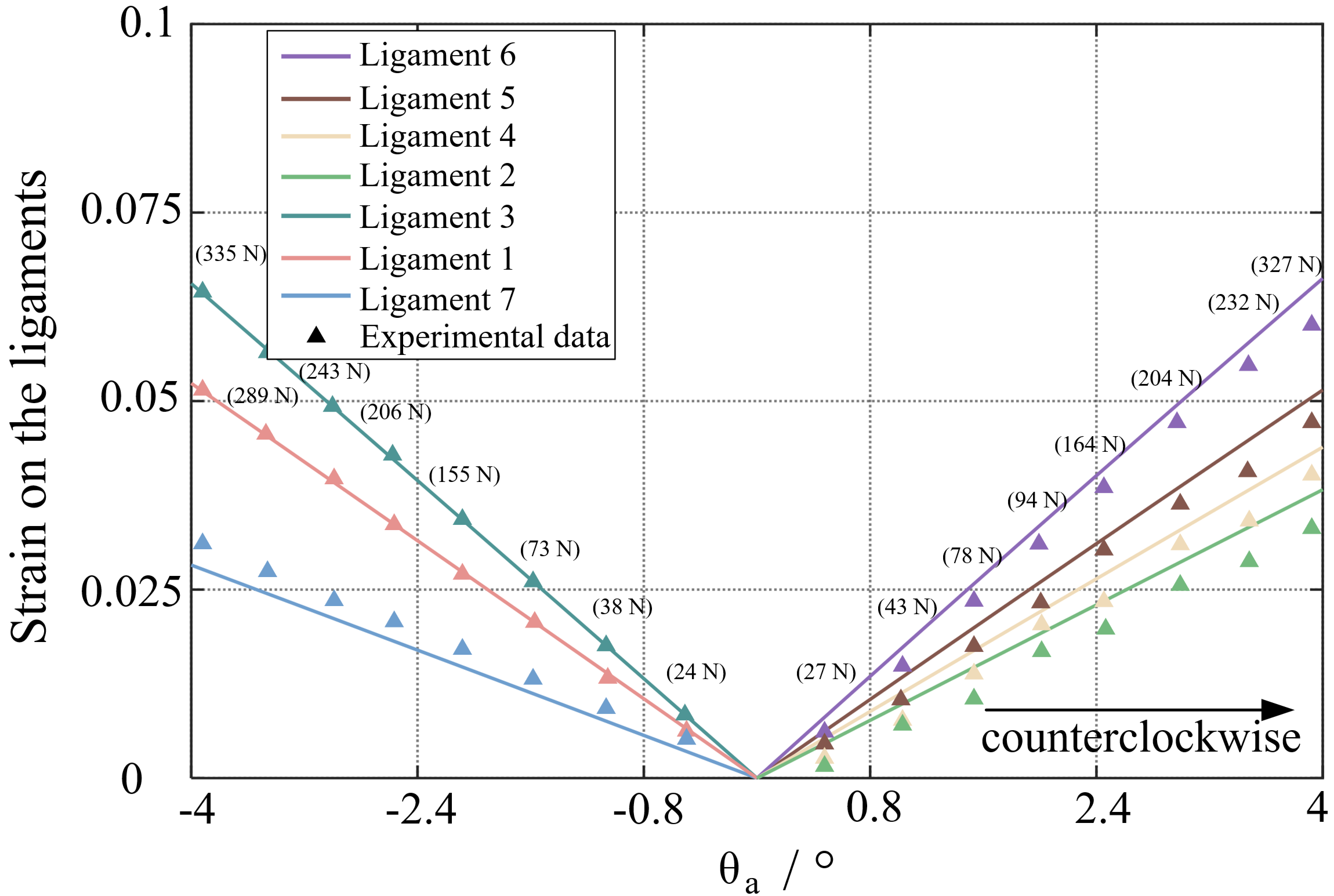}}
\caption{Simulation and experimental results of the relation between strain on each ligament of IOM and $\theta_a$. 
The enclosed values represent the magnitudes of lateral test forces applied on the distal forearm during the experiment.}
\label{fig6.13}
\end{figure}

For other ligaments in the IOM, the position parameters (listed in Table \ref{tab6.2}) can be substituted into the above calculation. These parameters include the position of the insertion points on the radius $l_g$ ($\overline{AG}$ or $\overline{AN}$), $\angle DAG$ (or $\angle DAN$), on the ulna $l_m$ ($\overline{BF}$ or $\overline{BM}$), $\angle CBF$ (or $\angle CBM$), and the initial length of the ligaments $l_f$ or $l_m$ ($\overline{FG}$ or $\overline{MN}$). The relationship between strain (the length change of $\overline{FG}$ or $\overline{MN}$) and $\theta_a$ for each ligament can be obtained, as illustrated in Fig. \ref{fig6.13}. It is evident that when the forearm undergoes lateral deflection, the strain on IOM ligaments increases, providing resistance to the lateral deflection of the forearm.

\begin{table}[htb]
\caption{The position parameter of ligaments in IOM.}
\footnotesize
\begin{center}
\begin{tabular}{c c c c c c c}
\toprule
& No. & $\bar{AG}$  & $\bar{BF}$ & $\bar{FG}$ & $\angle DAG$ & $\angle CBF$ \\
&  & or $\bar{AN}$ & or $\bar{BM}$ & or $\bar{MN}$ & or $\angle DAN$ & or $\angle CBM$\\
&  & (mm) & (mm) & (mm) & (\degree) & (\degree) \\
\midrule
$\circlearrowleft$ & 7 & 58.43 & 47.58 & 10.91 & -1.02 & 1.19 \\

 & 3 & 18.51 & 14.29 & 4.53 & -3.07 & 7.12\\

 & 1 & 11.56 & 5.83  & 6.01 & -7.14 & 17.64 \\
\midrule
$\circlearrowright$ & 6 & 38.58 & 42.77 & 4.78& 0     & 1.83  \\
 & 5 & 32.32 & 38.31 & 6.32 & 0 & 2.54 \\

 & 4 & 25.49 & 32.94 & 7.48 & -1.70 & 3.72 \\

 & 2 & 14.53 & 23.14 & 8.51 & -5.49 & 7.13 \\

\bottomrule
\end{tabular}
\label{tab6.2}
\end{center}
\end{table}

\begin{figure}[htb]
\centerline{\includegraphics[width=0.4\textwidth]{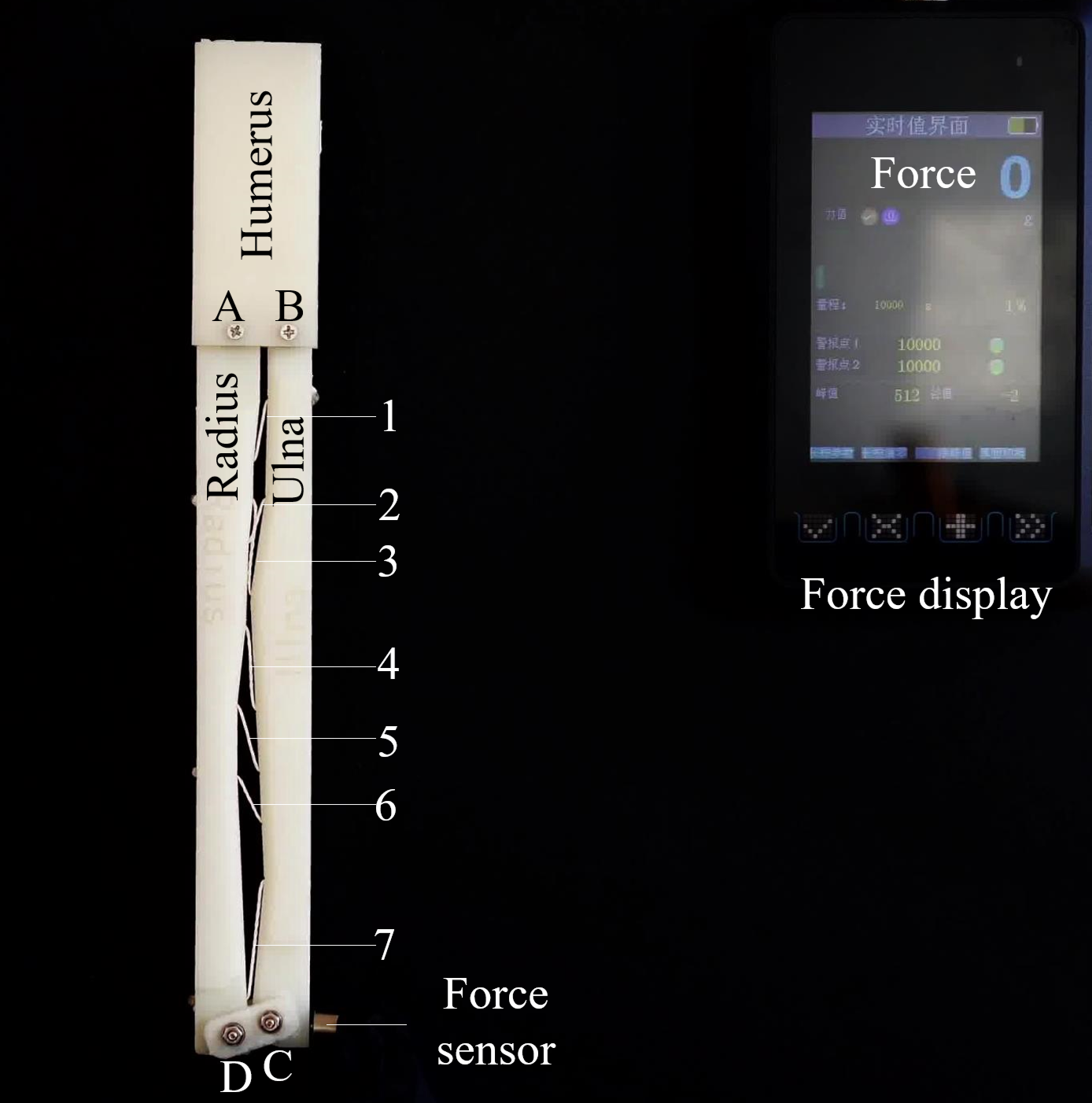}}
\caption{Test rig setup for validation of the relation between strain on each ligament of IOM and $\theta_a$.}
\label{fig7.12}
\end{figure}

The experimental setup, depicted in Fig. \ref{fig7.12}, was employed to validate the simulation results. The positioning of the IOM bundle insertion points on the radius ulna corresponded with the simulation data. The humerus was kept stationary, while the radius and ulna were hinged to the humerus at points A and B, thus enabling rotation within the delineated plane under the influence of lateral forces. This experimental arrangement mirrored the schematic outlined in Fig. \ref{fig6.12}(b), with the TFCC hinge-connected to the radius ulna at points D and C. The lateral force can be recorded by the force sensor (DYHW-108, measuring range: 10kg, accuracy: 0.3\%). The experimental procedures are demonstrably captured in Videos 1.1 and 1.2 in the supplementary videos.

In Figure \ref{fig6.13}, triangles denote experimental results, with bracketed force values indicating the magnitude of the test force causing forearm deflection. Measurements of IOM length variance and forearm deflection angle were taken using ImageJ software, with these results generally aligning with simulation results.


\subsection{Variation in MCL strain during elbow movement}\label{section8.0}

As the elbow is flexed or extended, approaching its range of motion limits, the strain in the MCL increases. This strain generates a pulling force that presses the ulna against the trochlea of the humerus (indicated by the red arrow in Fig. \ref{fig5.0}(d)). Through the annular ligament, IOM, and TFCC structures, the ulna exerts a pulling force on the radius (depicted by the blue arrow in Fig. \ref{fig5.0}(d)), enhancing the stability of the ball-and-socket joint by drawing the radius towards the capitulum of the humerus.

\begin{figure}[htb]
\centerline{\includegraphics[width=0.68\textwidth]{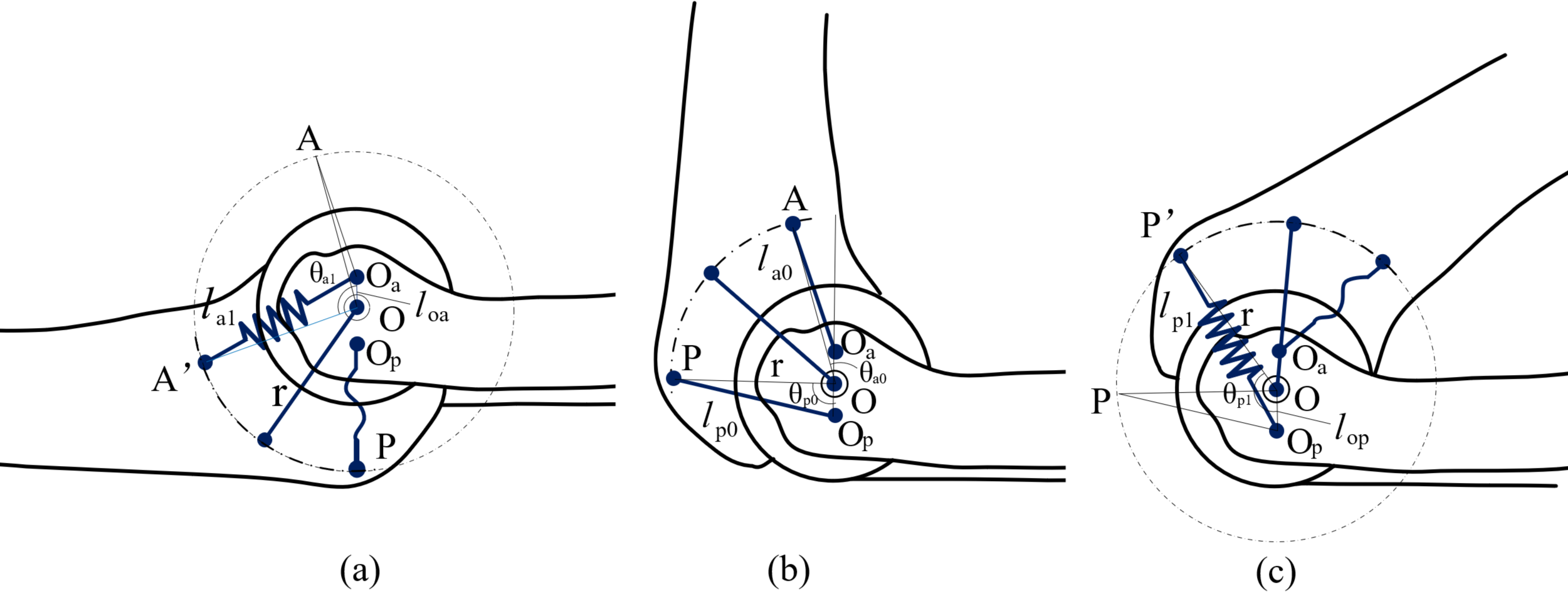}}
\caption{Simplified diagram of the MCL when the elbow is (a) fully extended, (b) at 90\degree and (c) at 135\degree.}
\label{fig6.15}
\end{figure}

Figure \ref{fig6.15} displays the length changes of the three components of the MCL as the elbow joint angle fluctuates. The anterior part and posterior part are denoted by lines $OA$ and $OP$, respectively. Composed of high-strength fibres, the ligaments exhibit spring-like characteristics. At the initial position of the elbow joint, with $\theta_{21}=90\degree$, all three components maintain their original lengths (Fig. \ref{fig6.15}(b)). The origin of the anterior part lies above the elbow rotation centre, at an eccentricity distance of $l_{oa}$ ($OO_{a}$). As the elbow extends (i.e., $\theta_{21}$ \textless $90\degree$), the anterior part lengthens from $l_{a0}$ to $l_{a1}$, transitioning from Fig. \ref{fig6.15}(b) to (a). The middle component's origin is situated at the rotation centre, maintaining a constant length. The posterior part's origin is positioned below the rotation centre, with an eccentricity distance of $l_{op}$ ($OO_{p}$). As the elbow flexes (i.e., $\theta_{21}$ \textgreater $90\degree$), the posterior part stretches from $l_{p0}$ to $l_{p1}$, as shown in the transition from Fig. \ref{fig6.15}(b) to (c).

According to the cosine law:
\begin{equation}
\begin{cases}
l_{a1}^2=l_{oa}^2+r^2-2l_{oa}r cos\theta_{a1} \\
l_{p1}^2=l_{op}^2+r^2-2l_{op}rcos\theta_{p1}
\end{cases}
\label{eq6.29}
\end{equation}

Where $\theta_{a1}$ denotes $\angle O_{a}OA'$ (Fig. \ref{fig6.15}(a)), $\theta_{a1}=\theta_{a0}+\pi/2-\theta_{21}$, and $\theta_{a0}$ represents $\angle O_{a}OA$ (Fig. \ref{fig6.15}(b)). $r$ refers to the length of $OP$. $\theta_{p1}$ is described as $\angle O_{p}OP'$ (Fig. \ref{fig6.15}(c)), with $\theta_{p1}=\theta_{p0}-\pi/2+\theta_{21}$ and $\theta_{p0}$ denoting $\angle O_{p}OP$ (Fig. \ref{fig6.15}(b)).

The strains in the anterior part $\varepsilon_a$, and posterior part $\varepsilon_p$ can be calculated as:
\begin{equation}
\begin{cases}
\varepsilon_a=(l_{a1}-l_{oa})/l_{oa} \\
\varepsilon_p=(l_{p1}-l_{op})/l_{op}
\end{cases}
\label{eq6.31}
\end{equation}

\begin{figure}[htb]
\centerline{\includegraphics[width=0.45\textwidth]{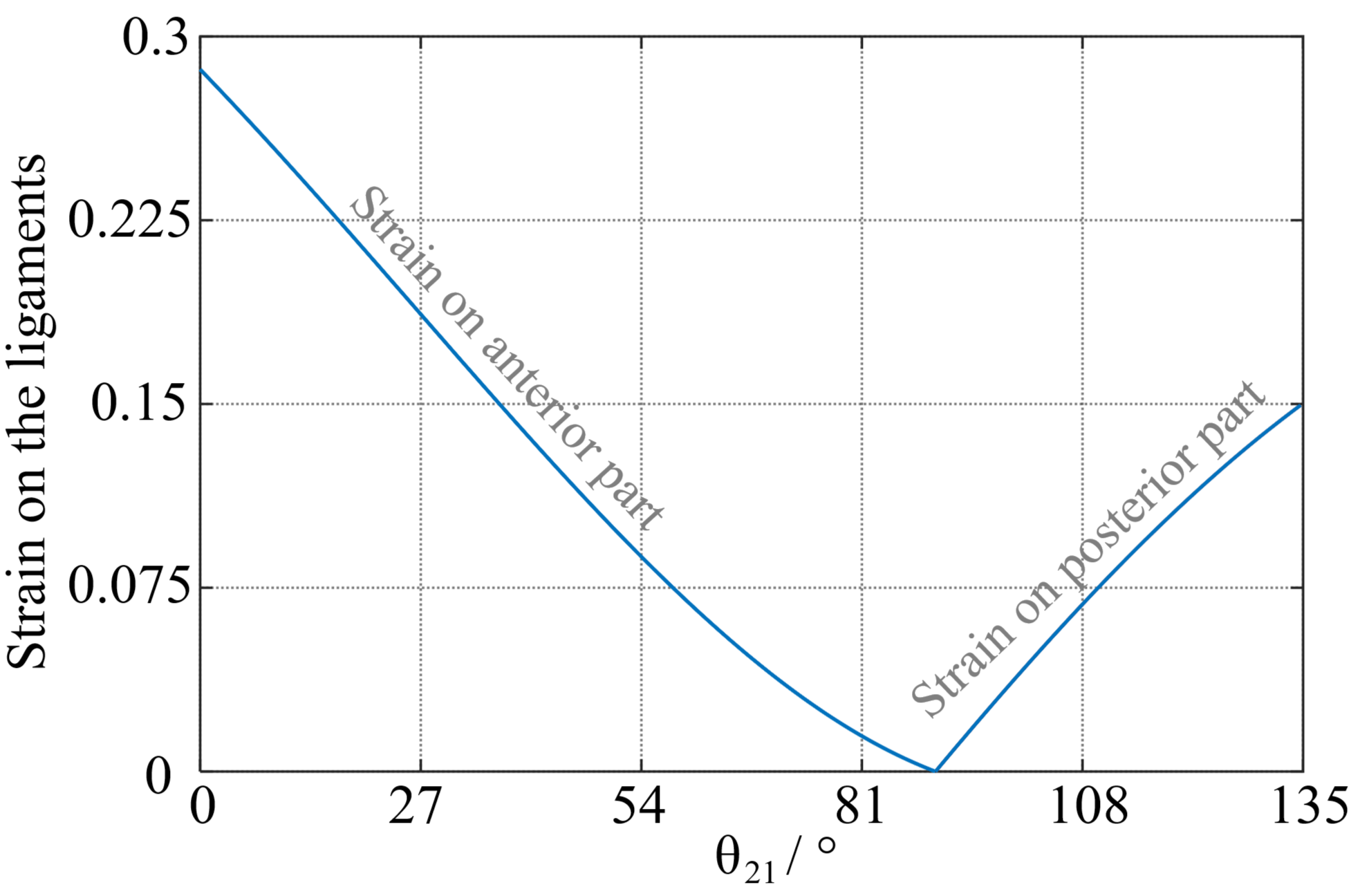}}
\caption{The simulation results of the strain on the MCL when angle of the elbow joint changes.}
\label{fig6.16}
\end{figure}

The variations in strain within the anterior and posterior portions as the elbow joint angle changes are shown in Fig. \ref{fig6.16}. As the strain in either the anterior or posterior part increases, a force is exerted on the ulna, resulting in the compression of the radius against the capitulum, further stabilizing the ball and socket joint between the radius and the humerus.

This section conducts a theoretical analysis of the mechanical intelligence discerned from the human arm and applies these principles to the proposed robotic arm design. The efficacy of these ingenious designs in enhancing arm performance remains to be determined. Consequently, the subsequent section will engage in constructing a physical prototype to validate the potential advantages of these designs.

\section{Skeleton-ligament Prototype and Muscular-skeleton actuation system}

In this section, a physical prototype is constructed based on the novel biomimetic robotic arm design delineated in the previous section. The primary focus is on replicating the soft tissues of the human forearm. Additionally, this subsection presents the actuation system and provides a computation of the output performance of the proposed robotic arm.

\subsection{Ligaments and adjustment mechanisms}

\begin{figure}[htb]
\centerline{\includegraphics[width=1\textwidth]{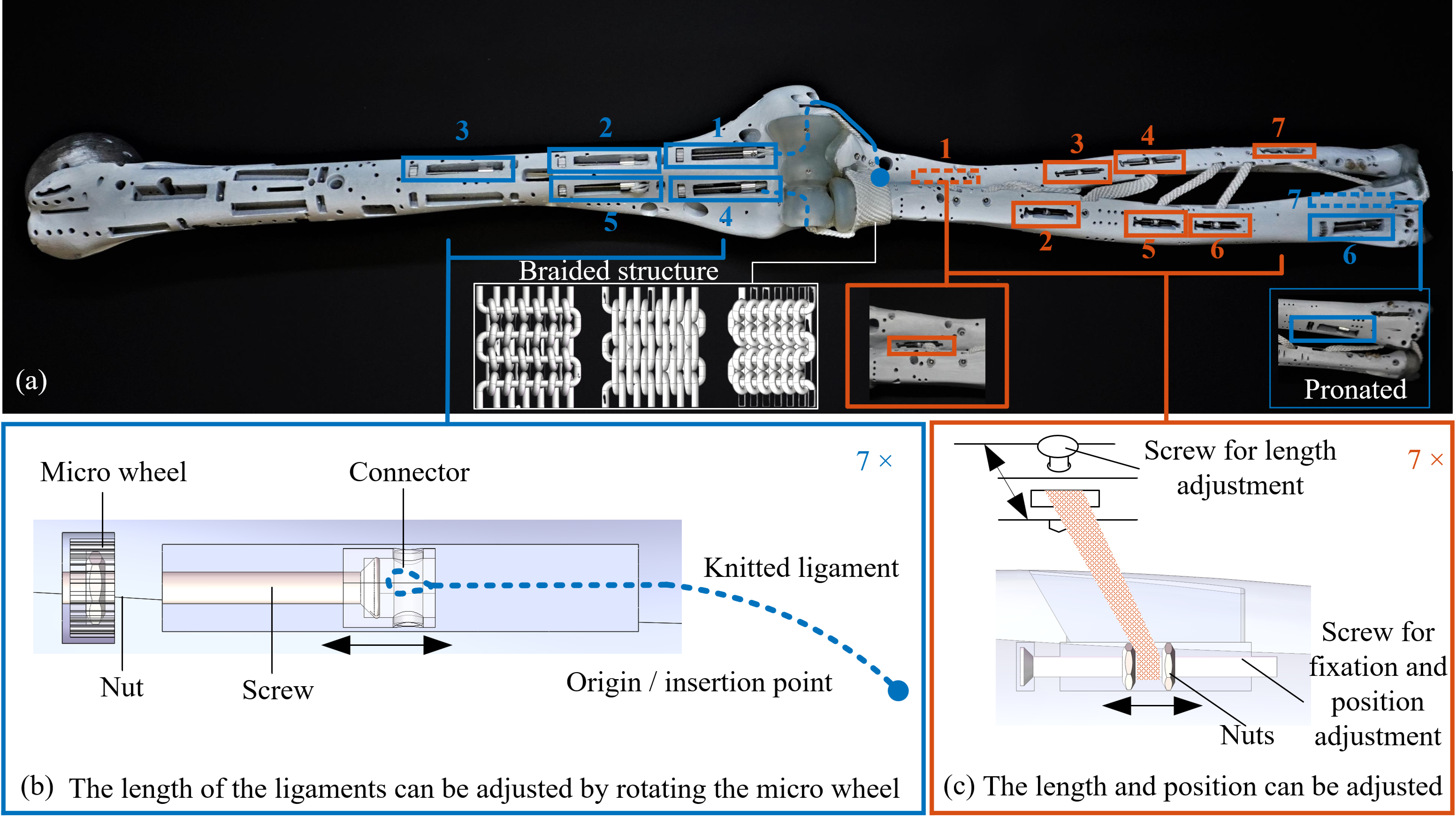}}
\caption{ligaments position/length adjustment mechanisms. (a) Overview of the 13 adjustment mechanism; (b) The length adjustment mechanism for ones marked in blue 1-7; (c) The position/length adjustment mechanism for ones marked in orange 1-7.}
\label{fig5.4}
\end{figure}

The ligament system is crucial for joint stabilization and restricting the range of motion. In the development of robotic forearms and elbows, ligaments exhibit a variety of shapes, sizes, and functions. For example, the annular ligament encircles the proximal head of the radius with a specific width. To increase strength and accommodate diverse shapes and sizes, ligaments are fabricated by intertwining multiple fibres, emulating the musculoskeletal system. Fig. \ref{fig5.4}(a) demonstrates an example of a braided structure created by interweaving seven fibres into a 2D configuration.

\begin{figure*}[htb]
\centerline{\includegraphics[width=1\textwidth]{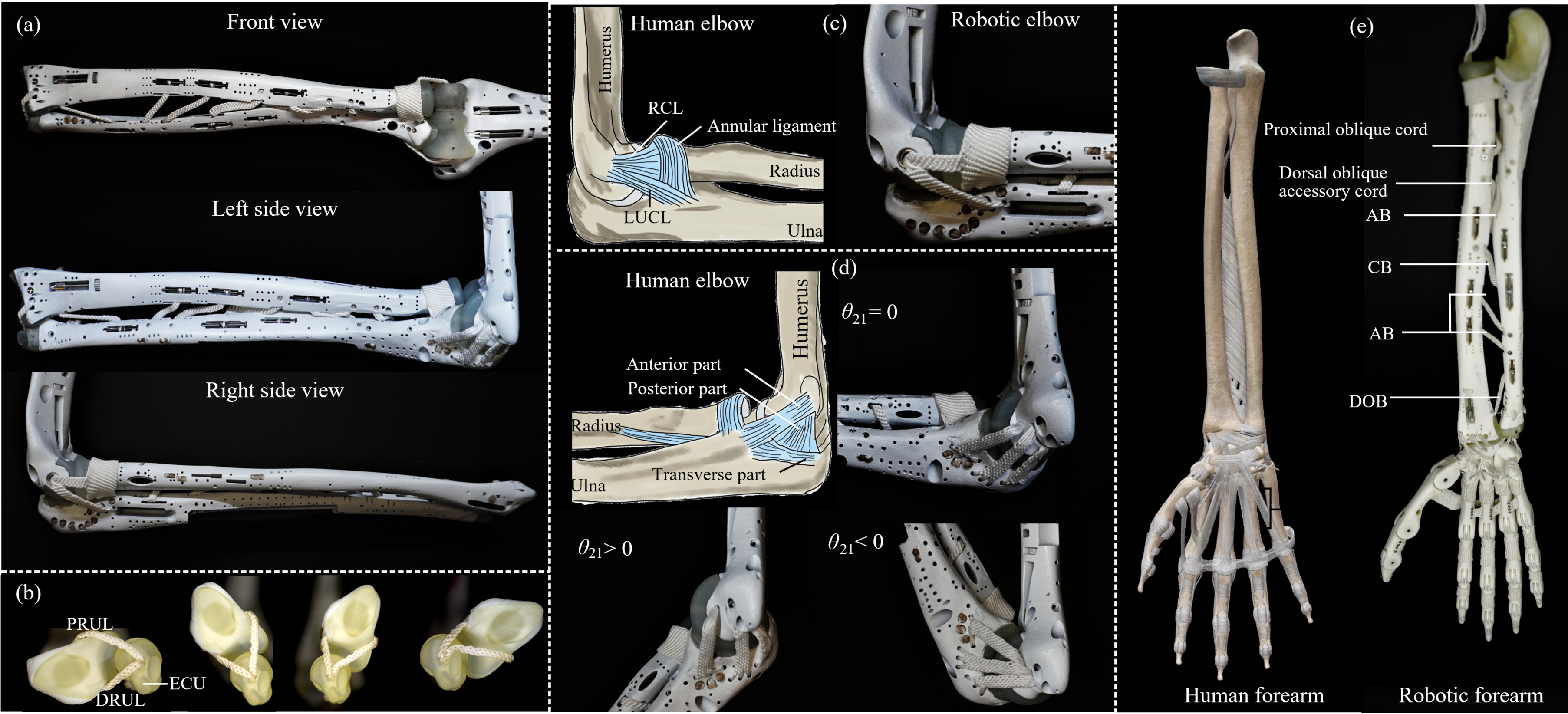}}
\caption{(a) The prototype of the skeleton-ligament system of the forearm and elbow, including front view, left and right side views; (b) Triangular fibrocartilage complex (TFCC) during pronation. The DRUL will slide across the distal ulna head and then across the ECU before starting to bend; (c) LCL including RCL and LUCL, annular ligament of the human and robotic elbow prototype; (d) The MCL of the human and robotic elbow; (e) The human forearm and hand, and the prototype of the robotic forearm and hand.}
\label{fig5.7}
\end{figure*}

To ensure the effective restrictive function of the ligaments, their lengths must be appropriately adjusted. For the MCL and LCL, five length adjustment mechanisms are employed, as illustrated in Fig. \ref{fig5.4}(a) and labeled as 1-5 in blue. The ligament bundles pass through the humerus' internal tubing and connect to the connectors in the adjustment mechanism, as depicted in Fig. \ref{fig5.4}(b). Rotating the micro wheel to manipulate the internal nut allows the adjustment screws to move axially within the slots, tightening or loosening the attached ligaments. This process enables the MCL and LCL to be adjusted to the optimal length for efficient functionality.

For the TFCC, mechanisms labeled as 6-7 (illustrated in blue) are deployed on the radius, as depicted in Fig.\ref{fig5.4}(a), with mechanism 7 situated at the rear of the radius. The underlying principle is congruent with that presented in Fig.\ref{fig5.4}(b). Modulating the length of the TFCC facilitates a 'soft-feel-end', {a condition where resistance escalates markedly as motion nears its limiting angle, contrasting with an abrupt halt due to rigid structures, during the rotational extremities of the forearm.} Specifically, the Distal RadioUlnar Ligament (DRUL) is modulated to maintain tension upon interfacing with the Extensor Carpi Ulnaris (ECU).

The position and length of each of the seven parts of the IOM can be adjusted using the embedded adjustment mechanism in the skeleton. Each adjustment mechanism's location is marked in orange box as 1-7 in Fig.\ref{fig5.4}(a). For instance, the No.2 adjustment mechanism is situated in the radius, as shown in Fig.\ref{fig5.4}(c). Adjusting the position of the nuts allows the length of the central band ligament (CB) to be modified in the axial direction up to 20mm. Adjusting the position and length of each portion is crucial to ensure that their inserted points are on the forearm rotation axis.

\subsection{Skeleton-ligament prototype and intelligent mechanisms}

The human musculoskeletal system serves as an ideal model for designing a robotic arm. To facilitate the design process, a 3D scanned model of human skeletons was optimized and utilized. The skeletons are printed with aluminium using SLM 3D printing technology, due to the low density and high strength of the aluminium. The articular cartilage, a thin and dense connective tissue covering joint surfaces, plays a crucial role in guaranteeing smooth joint contact and minimizing friction and wear during joint movements. To mimic the properties of articular cartilage, Formlabs durable resin is applied due to its durability, smoothness, and flexibility. As shown in Fig. \ref{fig5.7}(a), the articular cartilage is mechanically installed and glued onto the skeletons between each joint. To ensure adequate strength, the cartilage's average thickness is set at 1.5 mm while preserving the skeleton's original surface characteristics. Following the installation of the ligaments and the adjustment of their lengths to optimal positions, a prototype of the robotic elbow and forearm is developed, which emulates the human skeletal ligament system, as illustrated in Fig.\ref{fig5.7}(a).

The intelligent mechanisms discussed earlier have been incorporated into the prototype. As shown in Fig.\ref{fig5.7}(b), the TFCC structure starts to bend upon DRUL making contact with the ECU, leading to a sharp increase in strain, which restricts the forearm's rotational range while maintaining tension. Fig.\ref{fig5.7}(c) showcases the LCL and annular ligaments of the human elbow joint, along with their replicated counterparts on the elbow prototype. Their synergistic action securely connects the radius head to the humerus and ulna, while the ball-and-socket structure between the radial head and capitulum considerably improves the joint's dislocation resistance. Fig.\ref{fig5.7}(d) shows the MCL ligament of the elbow prototype, separated into three segments. As the elbow joint rotates bidirectionally, the strain on the MCL intensifies, enhancing the stability of the humeroulnar joint and subsequently pressing the radial head into the capitulum to further stabilize the humeroradial joint. Fig.\ref{fig5.7}(e) demonstrates the IOM replication on the prototype, which comprises seven sections. The IOM helps to stabilize the radius when axial or lateral forces are applied to the distal forearm and stabilizes the forearm against radioulnar bowing or splaying by drawing the ulna and radius toward the interosseous space. The external force is distributed between DRUJ and PRUJ. A robotic hand is attached to the robotic forearm using 5 ligaments as shown in Fig.\ref{fig5.7}(e).

\subsection{Muscular-skeleton actuation system}

\begin{figure}[htb]
\centerline{\includegraphics[width=1\textwidth]{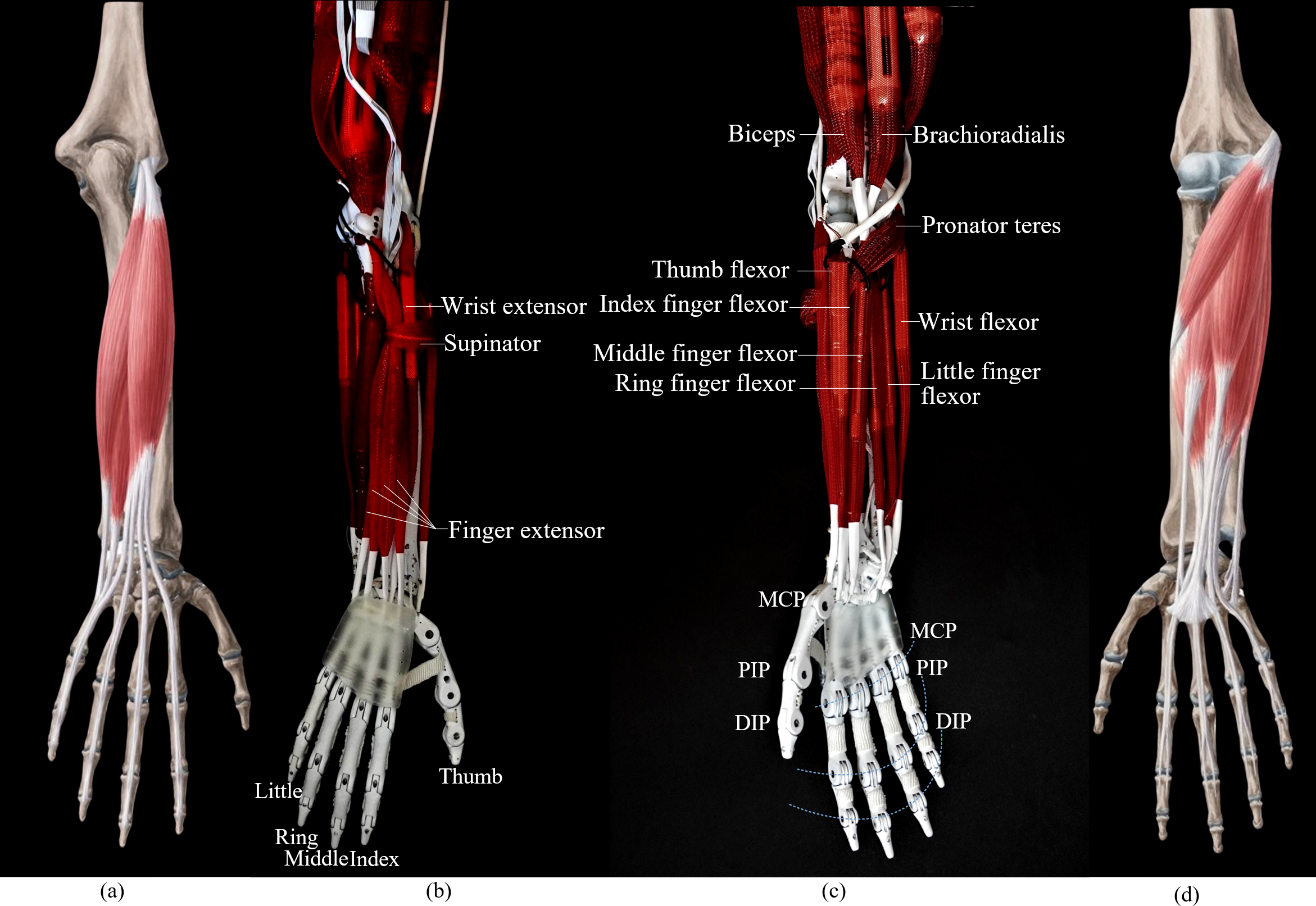}}
\caption{(a) Human forearm, posterior view; (b) Robotic forearm, posterior view; (c) Robotic forearm, anterior view; (d) Human forearm, anterior view.}
\label{fig9.10}
\end{figure}

To replicate the human elbow and forearm, the robot prototype is equipped with the biceps, brachioradialis, triceps, supinator, pronator teres,  and hand and wrist muscles, as shown in Fig. \ref{fig9.10}. For subsequent experiments, a simplified robotic hand is affixed to the robotic forearm using five ligaments, mirroring the skeletal system of the human hand. Each finger and thumb of the robotic hand is actuated by a pair of antagonistic muscles (flexor and extensor). Consistent with the human hand, all hand muscles are attached to the forearm.

\subsubsection{Elbow flexion/extension}

\begin{figure}[htb]
\centerline{\includegraphics[width=0.75\textwidth]{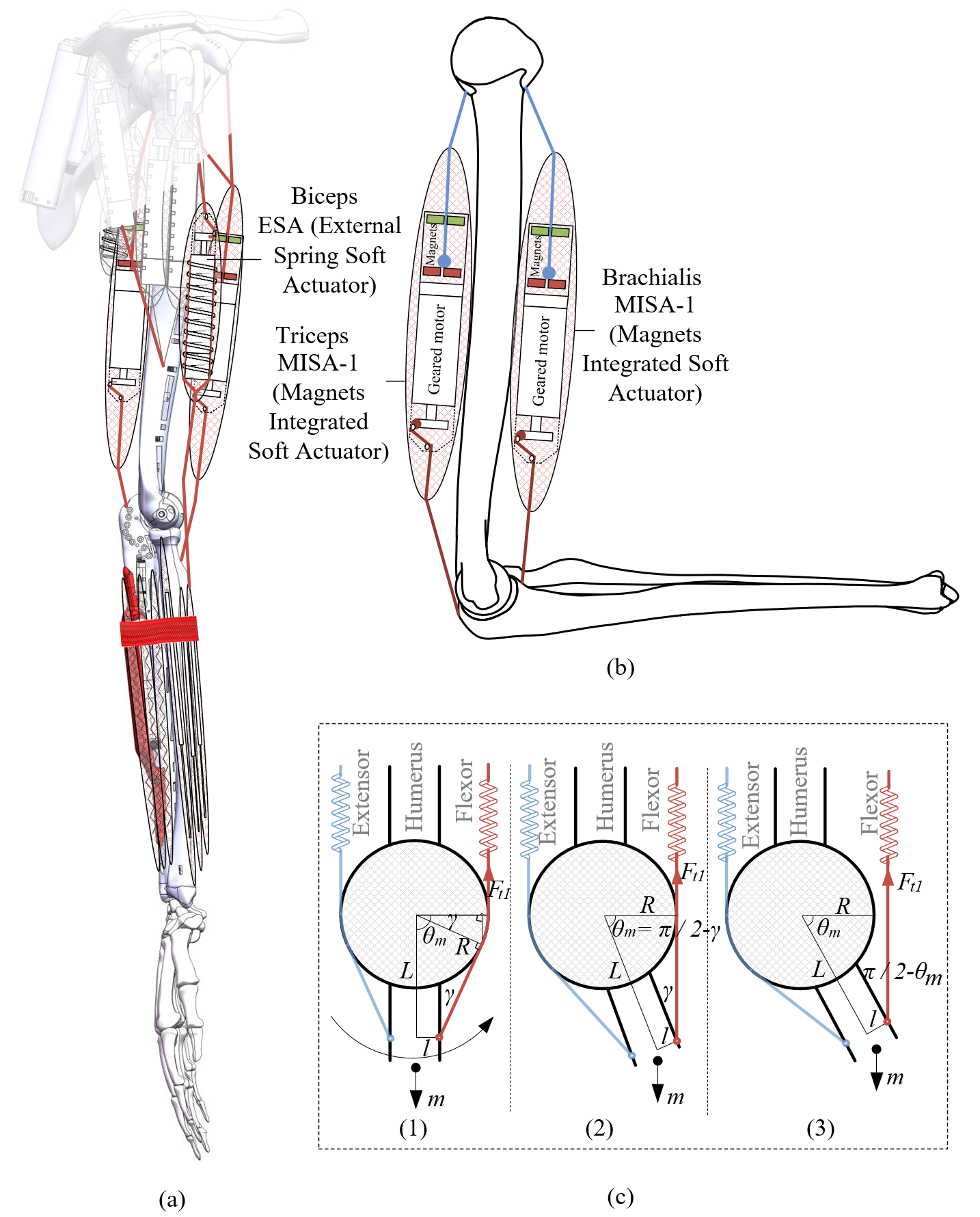}}
\caption{(a) The muscle arrangement of the elbow flexion/extension. (b) The brachialis and triceps assist in elbow rotation. (c) The simplified diagram of the elbow flexion actuation system in different $\theta_{21}$.}
\label{fig9.15}
\end{figure}

In this biomimetic arm prototype, both the brachialis and biceps contribute to elbow flexion. Drawing from our previous work\cite{yang2023novel}, a magnet-integrated soft actuator (MISA) with non-linear stiffness serves as the brachialis, originating from the humerus and connecting to the ulna. An external spring soft actuator with constant stiffness functions as the biceps, originating from the humerus and connecting to the radius. Another MISA operates as the triceps, originating from the humerus and connecting to the ulna, aiding in elbow extension. The actuation system configuration is shown in Fig. \ref{fig9.15}(a). As discussed in \cite{yang2023novel}, by utilizing two MISAs in an antagonistic configuration, the joint can achieve variable stiffness, effectively emulating the state of human joints as muscles tense and relax.

In daily life, elbow flexion often requires the ability to output substantial joint torque for performing everyday tasks. As illustrated in Fig. \ref{fig9.15}(c), when the flexor maintains the maximum output force $F_{t1}$ and the extensor only maintains tension (no force output), the joint torque $\tau_{21f}$ (flexion) varies with the joint angle $\theta_{21}=\pi/2-\theta_m$ ($\theta_{21}$ denotes the joint position, $\theta_m$ is illustrated in Fig. \ref{fig9.15}(c)) due to the moment arm's variation. There are three stages, as shown in Fig. \ref{fig9.15}(c), during which $\tau_{21f}$ can be calculated as:
\begin{equation}
{\tau}_{21f}={F}_{t1}(MR+Nr+PL)
\label{eq7.1}
\end{equation}

At stage 1 ($\theta_{m}>\pi/2-\gamma$), $M=1$, $N=0$, $P=0$; at stage 2 ($\theta_{m}=\pi/2-\gamma$), $M=0$, $N=cos\gamma$, $P=sin\gamma$; at stage 3 ($\theta_{m}<\pi/2-\gamma$), $M=0$, $N=sin\theta_{21}$, $P=cos\theta_{m}$. Where $\gamma=arcsin(R/\sqrt{L^{2}+l^{2}})-arctan(l/L)$.

\begin{figure}[htb]
\centerline{\includegraphics[width=0.55\textwidth]{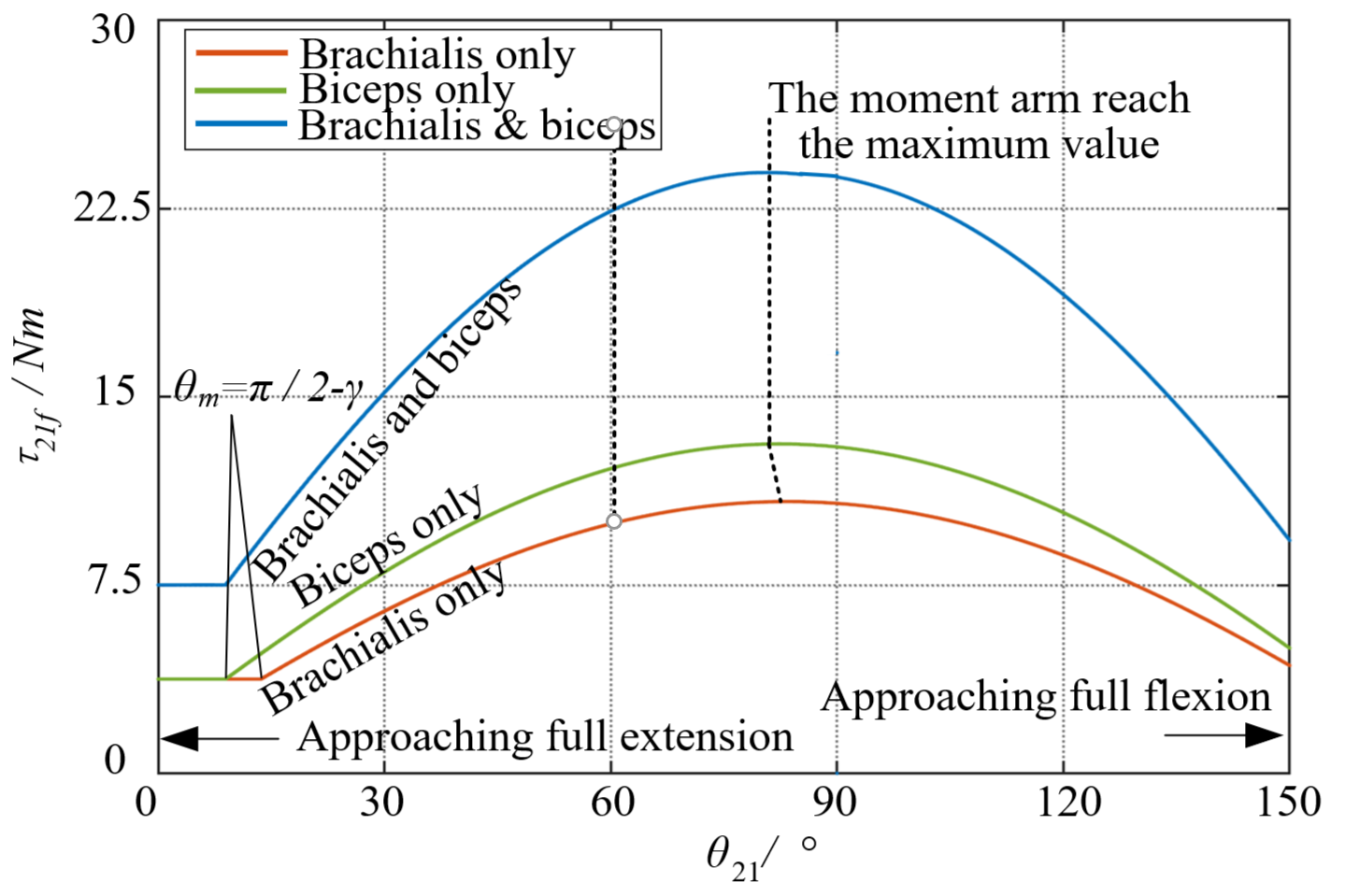}}
\caption{The simulation results of the relation between $\tau_{21f}$ and $\theta_{21}$.}
\label{fig7.2}
\end{figure}

Given the application of synthetic muscles (brachialis and biceps), the output force of the muscles are represented as: $F_{t1}$ = 250 N. Figure \ref{fig7.2} illustrates the simulation results for joint torque of elbow joint flexion as the joint angle varies while the actuator output force remains at its maximum. The red curve shows the joint torque driven by the brachialis alone, the green curve represents the biceps alone, and the blue curve represents both actuators working simultaneously. The results indicate that the output joint torque decreases as the elbow joint approaches full extension and full flexion, which is consistent with human joint behaviour. The results shows the peak torque for elbow flexion excess 24 Nm.

The elbow extension is actuated by the triceps (medial head, the output force is 250 N), and the joint torque $\tau_{21e}$ is constant (11.25Nm) as the joint angle $\theta_{21}$ changes.

\subsubsection{Forearm pronation}

\begin{figure}[htb]
\centerline{\includegraphics[width=0.75\textwidth]{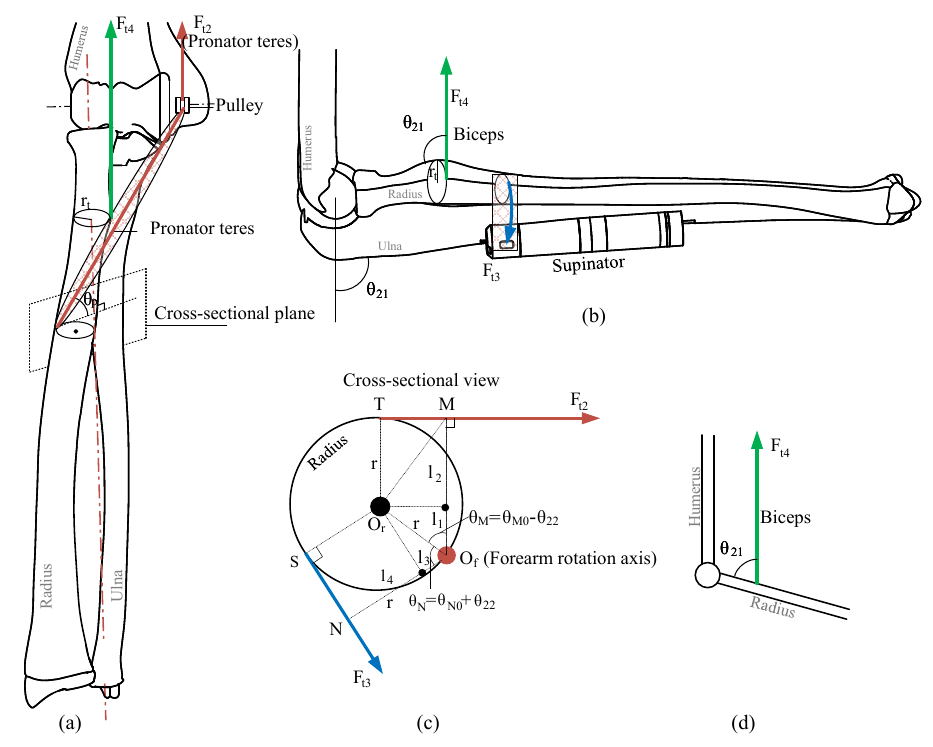}}
\caption{The actuation system of forearm pronation/supination: (a) front view, (b) side view; (c) cross-section view; (d) simplified diagram of the pronation actuation by biceps.}
\label{fig7.3}
\end{figure}

In the proposed design, forearm pronation is actuated by the pronator teres. As shown in Fig. \ref{fig7.3}(a), the motor for the pronator teres is located on the side of the humerus, with a pulley fitted inside the humerus to minimize friction. The rotation axis of the pulley coincides with the axis of the humeroulnar joint. The red tendon passes through the pulley, extends across the ulna and radius, and is ultimately fixed to the lateral side of the radius. Notably, elbow flexion or extension does not affect the length of the pronator teres tendon.

When the pronator teres drives forearm pronation, the cross-sectional view of the structure in the plane perpendicular to the forearm rotation axis (Fig. \ref{fig7.3}(a)) can be simplified to Fig. \ref{fig7.3}(c). The red line represents the projection of the tendon, and the angle between the tendon and the sectional plane is denoted as $\theta_p$ (Fig. \ref{fig7.3}(a)). The tendon contacts the radius at point $T$ and exerts a force that rotates the radius around the forearm rotation centre, marked in red as $O_f$ in Fig. \ref{fig7.3}(c). The cross-sectional view of the radius can be approximated as a circle with centre $O_r$ and radius $r$, which passes through $O_f$. As the radius rotates, $\theta_M=\theta_{M0}-\theta_{22}$ decreases, where $\theta_M$ represents $\angle MO_f O_r$, $\theta_{M0}$ is the initial value of $\theta_M$, and $\theta_{22}$ is the radius rotation angle. The output torque $\tau_{22p}$ during forearm pronation can be calculated as follows:

\begin{equation}
\tau_{22p}=F_{t2}(l_1+l_2)
\label{eq7.2}
\end{equation}

Where $l_1=r cos(\theta_{M0}-\theta_{22})$ and $l_2=r$. $F_{t2}$ is the tendon force. 

\begin{figure}[htb]
\centerline{\includegraphics[width=0.68\textwidth]{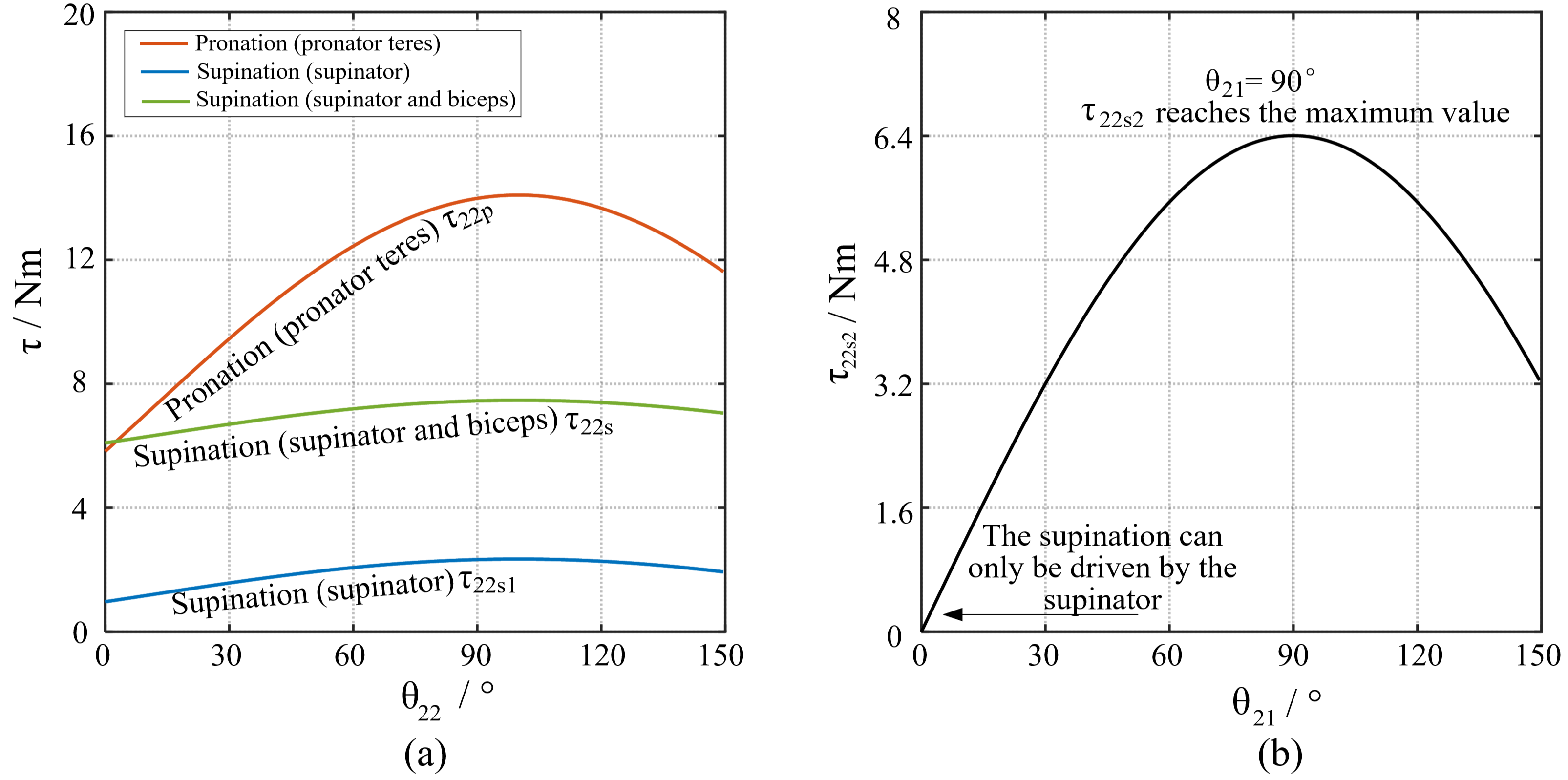}}
\caption{(a) The simulation result of the relation between $\tau_{22p}$, $\tau_{22s1}$, $\tau_{22s}$ and $\theta_{22}$. (b) The simulation result of the relation between $\tau_{22s2}$ and $\theta_{21}$.}
\label{fig7.4}
\end{figure}

The relationship between the rated torque $\tau_{22p}$ and the forearm rotation angle $\theta_{22}$ is shown in red in Fig. \ref{fig7.4}(a) when the maximum output force of the motor is kept constant $F_{t2}$ = 734 N). It is noticeable that $\tau_{22p}$ attains the highest value when $\theta_{22}$ approaches $100\degree$, which is 14 Nm.

\subsubsection{Forearm supination}

The forearm supination is driven by both the supinator (marked in blue in Fig. \ref{fig7.3}(b)) and the biceps (marked in green in Fig. \ref{fig7.3}(a) and (b)). The motor of the supinator is installed inside the ulna to reduce the size of the actuator. The tendon of the supinator passes over the outer side of the radius and is fixed to the inner side of the radius, while the tendon of the biceps wraps around the radial head and attaches to the proximal end of the radius. It is worth noting that the insertion points of the supinator and the pronator teres on the radius are located on the same intercept plane perpendicular to the forearm rotation axis. This ensures that the supinator and pronator teres can be balanced during forearm rotation, providing a stable and smooth movement.

When the supinator drives the forearm supination, the section view of the structure can be simplified as in Fig. \ref{fig7.3}(c). The blue line is the projection of the tendon on the sectional plane. The tendon is contacted with the radius at point $S$, pulling the radius rotates around the forearm rotation centre $O_f$. $\theta_N=\theta_{N0}+\theta_{22}$ increases as the radius rotates, where $\theta_{N0}$ is the initial value of $\theta_N$. The torque when the supinator drives the forearm supination is:

\begin{equation}
\tau_{22s1}=F_{t3}(l_3+l_4)
\label{eq7.3}
\end{equation}

Where $l_3=rcos(\theta_{N0}+\theta_{22})$ and $l_4=r$. $F_{t3}$ is the tendon force from the motor.

Presuming the force values of $F_{t3}$ = 122 N, the correlation between $\tau_{22s1}$ and $\theta_{22}$ can be elucidated as depicted in Fig. \ref{fig7.4}(a), denoted by the blue marking.

The insertion point of the biceps muscle is located on the radial tuberosity. To illustrate the relationship between the biceps and the radius, a section view of the radial tuberosity on a plane perpendicular to the forearm rotation axis is shown in Fig. \ref{fig7.3}(b). This view depicts the radial tuberosity as a circle with a radius $r_t$. The torque produced by the biceps to drive forearm supination can be calculated as follows:

\begin{equation}
\tau_{22s2}=F_{t4}r_t\sin(\theta_{21})
\label{eq7.4}
\end{equation}

Where $F_{t4}$ is the tendon force of the biceps, $\theta_{21}$ is the angle between the biceps and the radius, as shown in Fig. \ref{fig7.3}(d).

The plot in Fig. \ref{fig7.4}(b) shows the relationship between $\tau_{22s2}$ and $\theta_{21}$ ($F_{t4}$ = 250 N). As $\theta_{21}$ approaches 0, the value of $\tau_{22s2}$ converges to 0, indicating that the forearm supination can only be driven by the supinator at this point. When $\theta_{21}=90\degree$, $\tau_{22s2}$ reaches its maximum value.

When $\theta_{21}=90\degree$, the joint torque when forearm supination, $\tau_{22s}=\tau_{22s1}+\tau_{22s2}$ in relation to $\theta_{22}$ as shown in Fig. \ref{fig7.4}(a) marked in green. $\tau_{22s}$ shows a small fluctuation during forearm rotation, with a maximum value close to 7.8 Nm.

\section{Validation}

This section investigates the Soft-Feel-End mechanism of the robotic arm by soft tissues and assesses the individual contributions of these soft tissues to joint stability. Ultimately, a demonstration of the motion performance of the proposed robotic arm is presented.

\subsection{Validation of Soft-Feel-End mechanism in regulating forearm positioning}

\begin{figure}[htb]
\centerline{\includegraphics[width=0.75\textwidth]{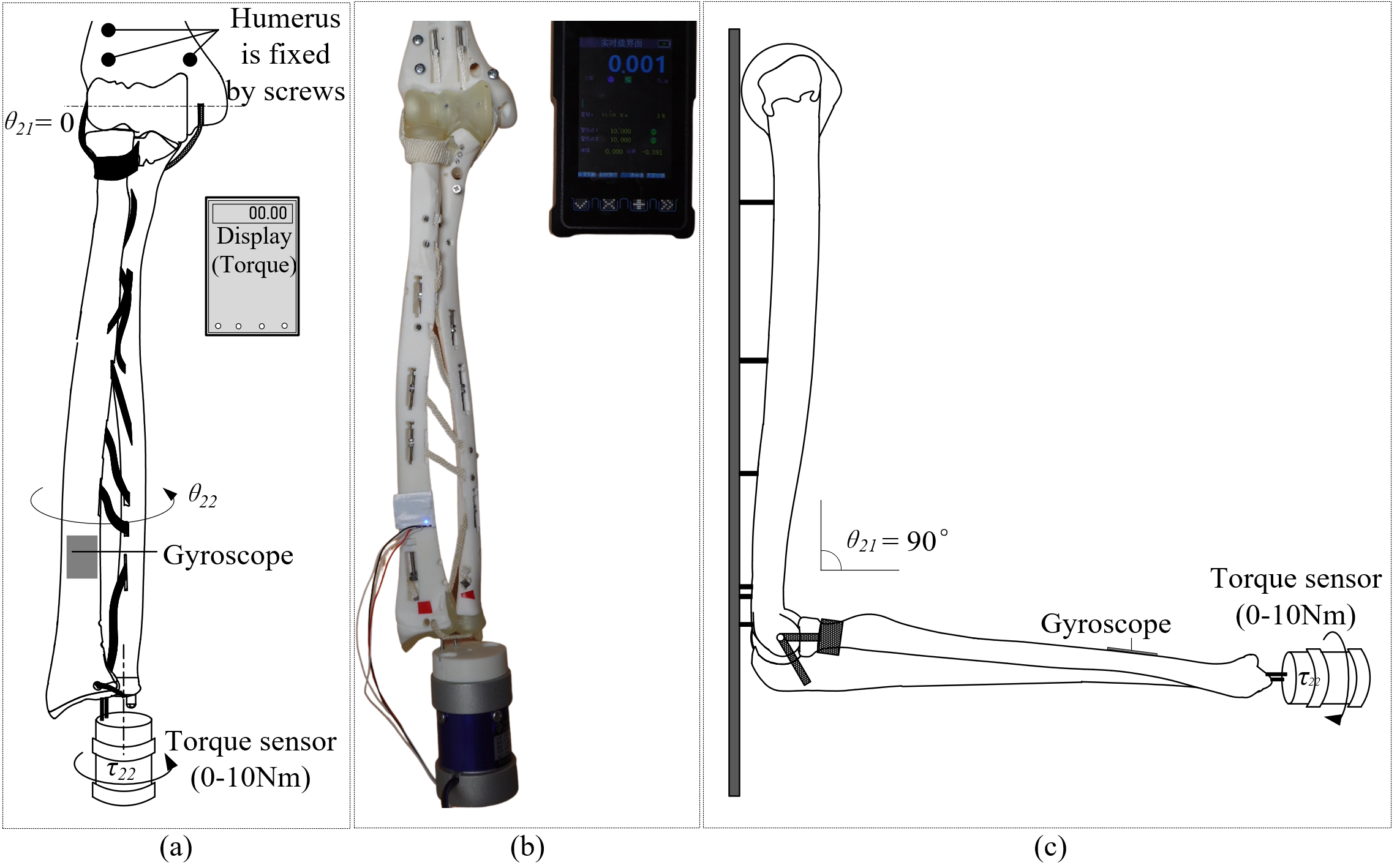}}
\caption{(a) The schematic diagram of the test rig for forearm rotation resistance measurement (front view); (b) Test rig setup for forearm rotation resistance measurement; (c) side view.}
\label{fig9.6}
\end{figure}

In the foregoing analysis, it was determined that as the forearm rotation approaches its limited position, the resistance increases rapidly, functioning as a position-limiting mechanism. In this experiment, the resistance torque during the forearm rotation of the proposed skeletal model will be measured by passively driving the forearm rotation at various elbow joint angles.

The experimental setup is shown in Fig. \ref{fig9.6}, where the humerus is secured to the base plate, and the ulna and radius remain unconstrained. The forearm prototype is positioned vertically. The initial forearm rotational position is set to $\theta_{22}=0$, corresponding to full supination. The elbow joint position is set to $\theta_{21}=0$, indicating full extension. The torque sensor {(Brand: Dayang sensor, Model: DYJN-104, Capacity: 0-10 Nm, Rated Output: 2.0 mV/V)} is attached to the radius using screws, and its rotation axis aligns with the forearm rotation axis. The gyroscope {(Brand: Wit-motion, Model: WT901BLECL, Chip: MPU9250)} records the rotation of the radius, while the torque sensor records the external torque required for rotating the radius.

The experiment steps are:

Step 1: Set the elbow joint to the initial position and maintain it.

Step 2: Rotate the torque sensor manually so that the radius rotates from fully supinated to fully pronated. Record the radius position and the external torque applied.

Step 3: Change $\theta_{22}$ and repeat the experiment.

\begin{figure}[htb]
\centerline{\includegraphics[width=0.65\textwidth]{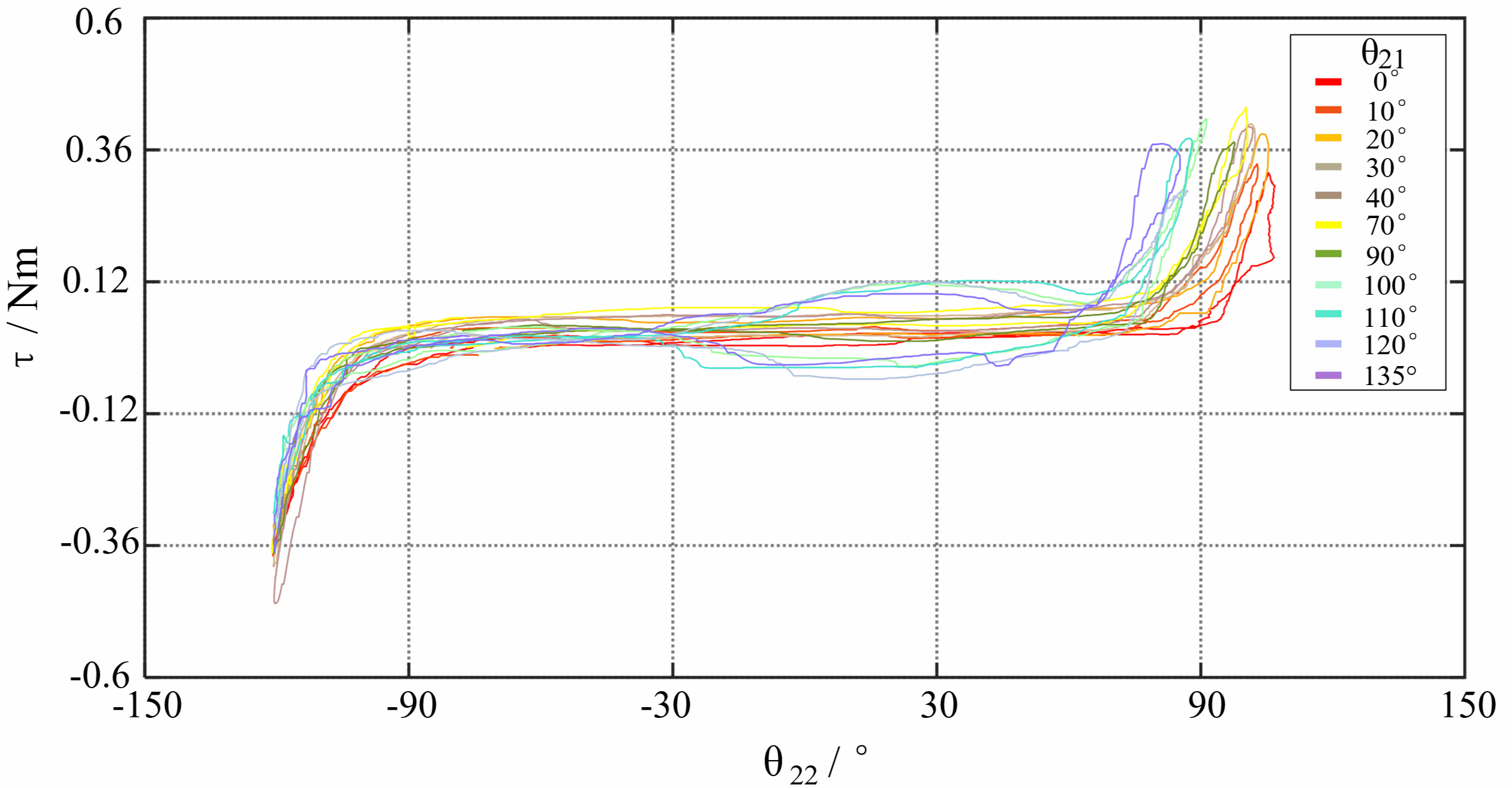}}
\caption{The experiment results of forearm rotation resistance measurement with different $\theta_{21}$.}
\label{fig9.7}
\end{figure}

The experimental results are presented in Fig. \ref{fig9.7}. It can be observed that, for any $\theta_{21}$, as $\theta_{22}$ approaches the limited position, the resistance torque increases, and the soft restriction is achieved. When $\theta_{21}$ exceeds 100\degree, indicating elbow joint flexion, the resistance torque displays a noticeable increase within the range of $-30\degree<\theta_{22}<60\degree$. This might be attributed to the tensioning of the posterior part of MCL when $\theta_{21}>90\degree$, causing the radius to be pressed into the humeroradial joint and increasing frictional resistance, as discussed in \ref{section8.0}. 

\subsection{Validating the contribution of IOM, TFCC, annular ligament to forearm stability}

In order to evaluate the individual contribution of the different parts of the IOM ligaments to the stability of the forearm, an experiment was conducted to measure the deflection angle of the forearm when subjected to lateral external forces with partial IOM ligaments disabled. The experimental setup is shown in Fig. \ref{fig9.3}. The forearm rotation angle is set to $\theta_{22}=0$, i.e., fully supinated, and remains in that position. The rotation of the radius was recorded by the gyroscope. The experiment steps are:

\begin{figure}[htb]
\centerline{\includegraphics[width=0.65\textwidth]{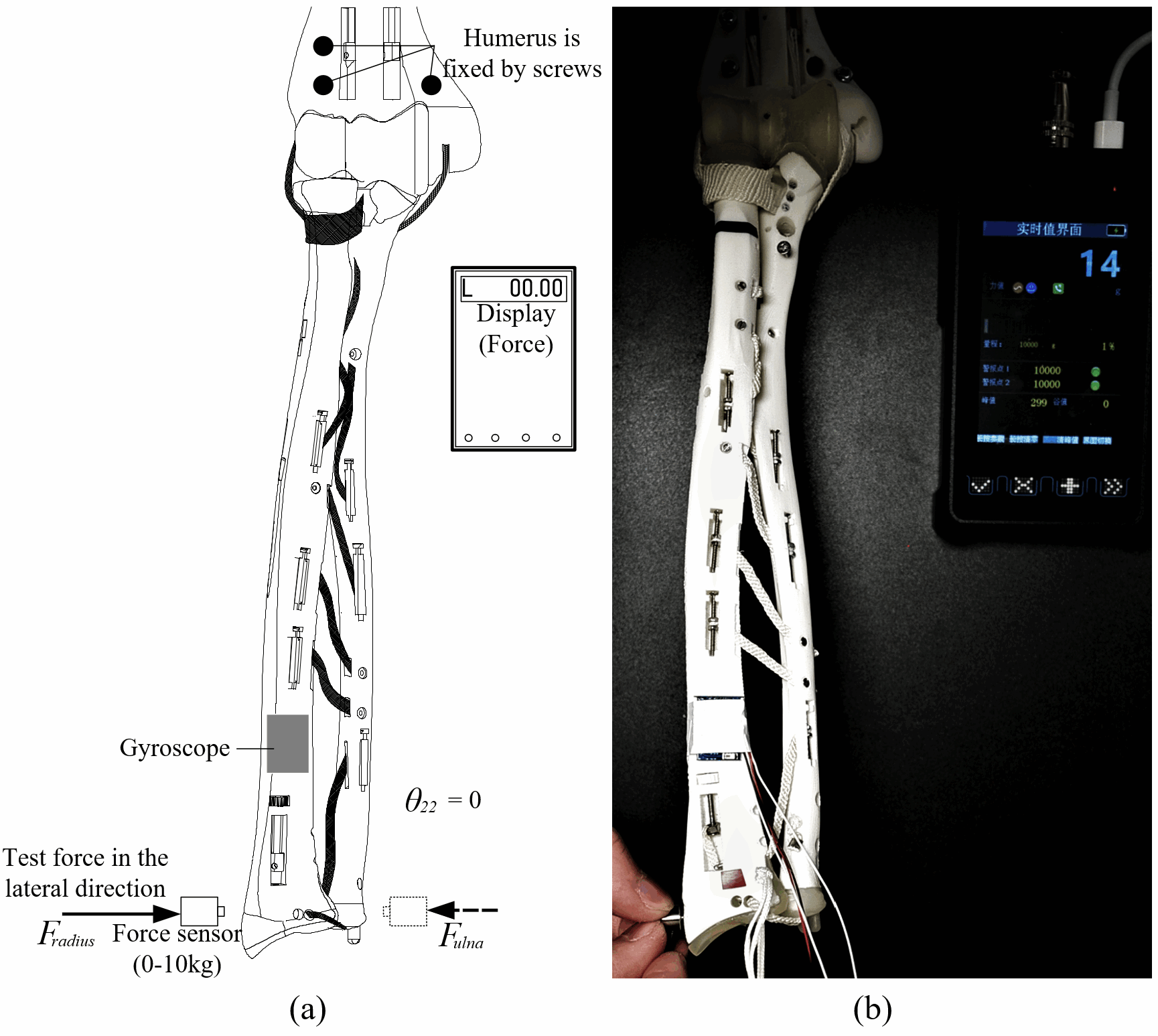}}
\caption{(a) The schematic diagram of the test rig for validation of the contribution of the soft tissue to forearm stability in a lateral direction; (b) Test rig setup for validation of the contribution of the soft tissues to forearm stability in the lateral direction.}
\label{fig9.3}
\end{figure}

Step 1: Keep all bundles of the IOM intact.

Step 2: Apply a lateral force $F_{radius}$ at the distal end of the radius, ensuring that the point of application and the maximum force is the same for each test. Record the external force and the deflection of the radius during the test.

Step 3: Apply a lateral force $F_{ulna}$ at the distal end of the ulna, keeping the distance between the application point of $F_{ulna}$ and $F_{radius}$ from the elbow rotation axis the same. The point of application and the maximum of $F_{radius}$ are kept the same for each experiment. Record the external force and the deflection of the radius during the test.

Step 4: Disable specific bundles of the IOM ligaments and repeat the experiment.

\begin{figure}[htb]
\centerline{\includegraphics[width=0.8\textwidth]{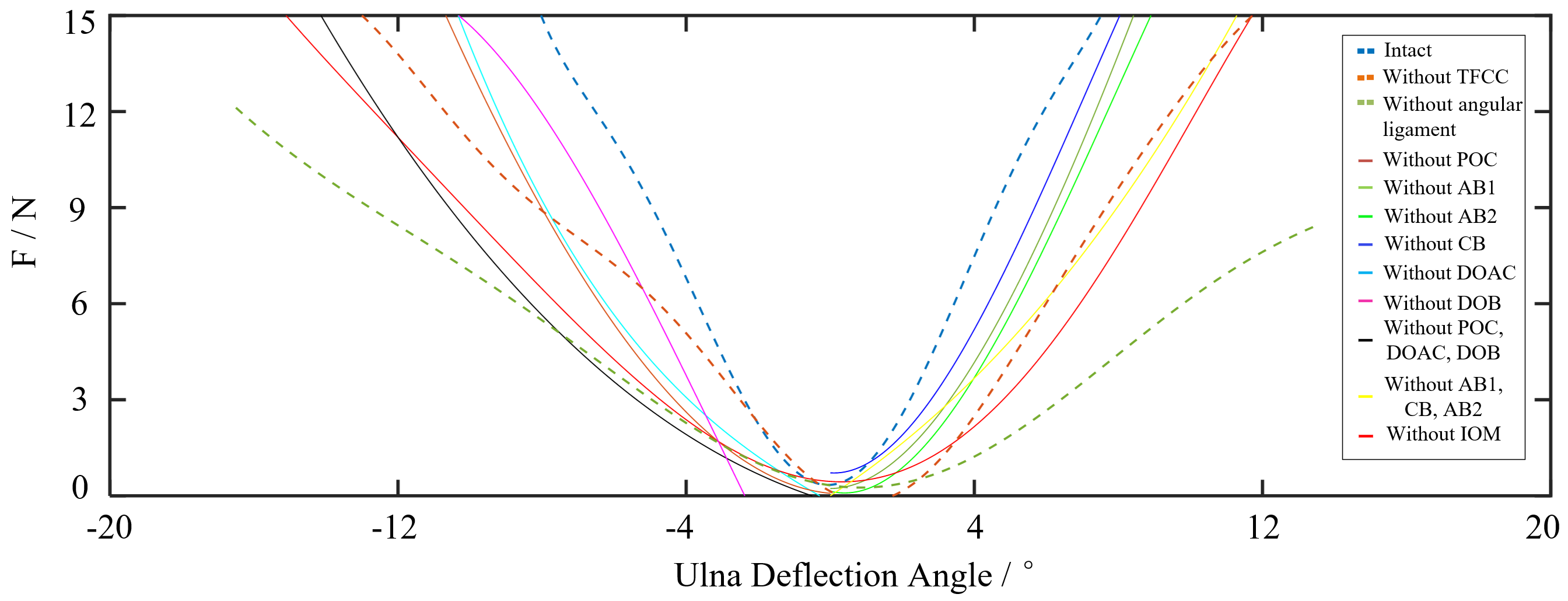}}
\caption{The experimental results of the test of the contribution of IOM, TFCC, angular ligament to forearm stability.}
\label{fig9.4}
\end{figure}

The experimental results presented in Fig. \ref{fig9.4} demonstrate that intact IOM ligaments contribute to enhanced stability in the forearm, as evidenced by the smallest deflection angle measured when all ligaments are intact. However, when certain ligament groups are disabled, such as POC, DOAC, and DOB, the ability of the forearm to resist lateral forces is weakened, resulting in larger deflection angles when the lateral force is applied to the left. Disabling AB and CB ligaments lead to an increase in the counterclockwise deflection angle. Upon partial absence of the IOM, the angular deflection of the forearm experienced a notable increase when subjected to a leftward lateral force ($F_{ulna}$ = 12N) as compared to the intact IOM scenario: 15.48\% (CB), 24.32\% (AB1), 30.1\% (AB2), 63.69\% (AB1, CB, AB2), and 72.19\% (IOM). Similarly, when the forearm was exposed to a rightward lateral force ($F_{radius}$ = 12N), the angular deflection experienced a marked increase: 24\% (DOB), 42.55\% (DOAC), 45.19\% (POC), 92.55\% (POC, DOAC, DOB), and 95.65\% (IOM). These results suggest that each IOM ligament group plays a significant role in stabilizing the forearm under lateral external forces.

Importantly, the experimental validation of the IOM ligament's function was performed using a forearm prototype printed with polylactic acid (PLA). Given the discrepancy in material strength, it's anticipated that the deflection angle observed in the experimental results would considerably exceed that of an equivalent forearm skeleton prototype fabricated from aluminium."

The experimental validation of the contribution of TFCC and annular ligament (when the IOM is intact) to the lateral stability of the forearm was also carried out using the test rig shown in Fig. \ref{fig9.3}. The results are presented in Fig. \ref{fig9.4}. {Without an annular ligament, applying a test force to the distal forearm and initiating a clockwise rotation causes the ulna and the radius to separate, leading to a complete disintegration of the forearm.} Conversely, applying the opposite test force and executing a counterclockwise rotation results in a 153.1\% increase in the deflection angle of the radius, compared to instances where the annular ligament is intact. The findings suggest that when both TFCC and annular ligament are intact, the deflection of the radius is minimal, and the forearm is stable. When the annular ligament is disabled, the deflection is the largest, indicating significant instability of the forearm.

\subsection{Validation of performance}

\begin{table*}[htbp]
\caption{Performance of robotic elbow and forearm.}
\footnotesize
\begin{center}
\begin{tabular}{l c c c c}
\toprule
\makecell[c]{Motion group} & \makecell[c]{Range of motion} & \makecell[c]{Percentage$^1$} & \makecell[c]{Joint torque (Nm)} & \makecell[c]{Percentage$^2$} \\
\midrule
Elbow Extension(-)/Flexion(+) & 0-140.25\degree & 98.8\% & -11.25 to 24 & 27.2\% / 33.7\%\\
Forearm Pronation(-)/Supination(+) & -60\degree-51.5\degree & 58.7\% & -14 to 7.8 & 195.5\% / 87.3\%\\
\bottomrule
\end{tabular}

 \begin{tablenotes}
        \footnotesize
        \centering
        \item[1] $^1$The percentage of motion ranges compared to biological joints. $^2$The percentage of joint toques compared to biological joints.        
      \end{tablenotes}
\label{tab2.9}
\end{center}
\end{table*}

\begin{figure}[htb]
\centerline{\includegraphics[width=0.6\textwidth]{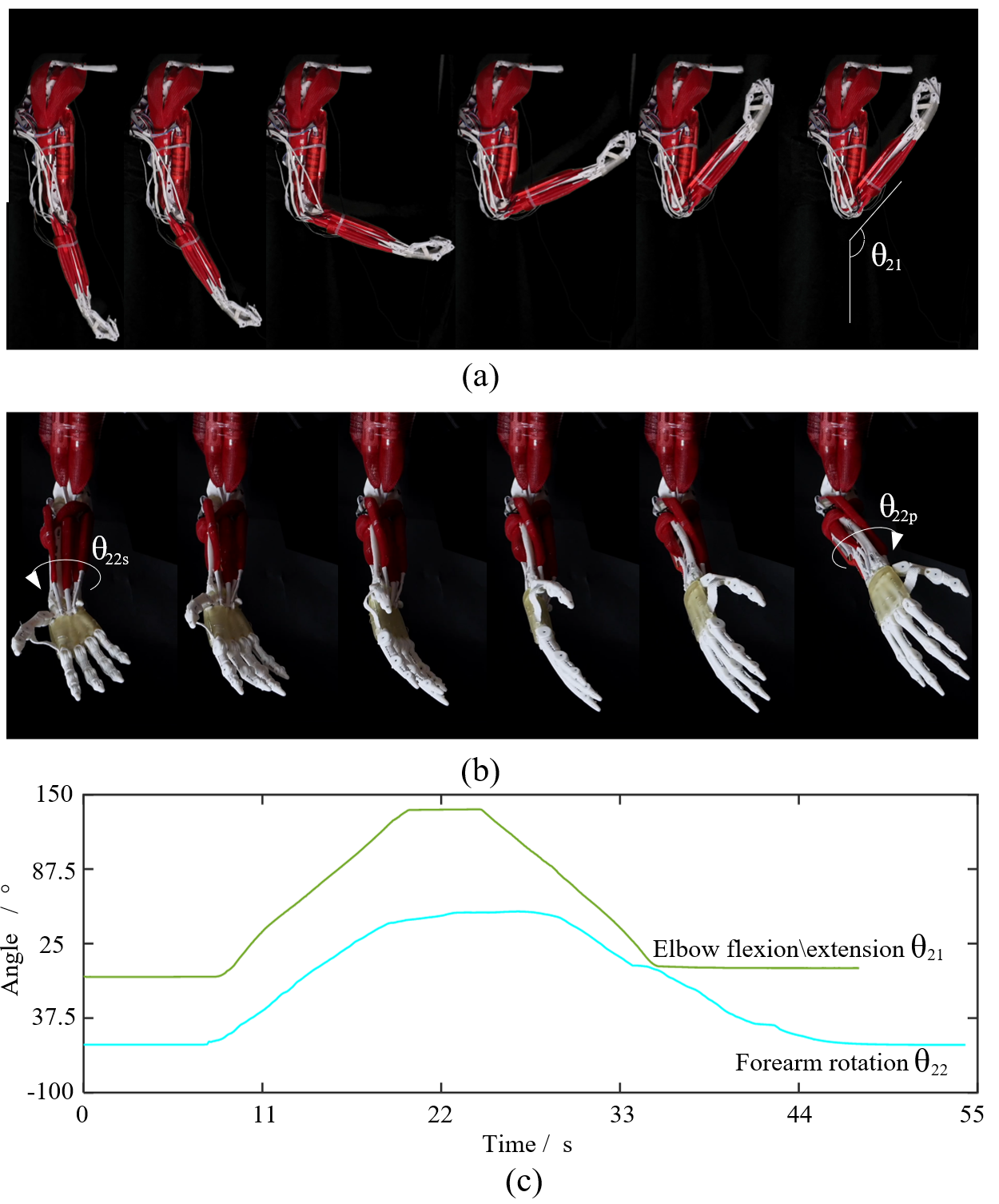}}
\caption{(a) Elbow flexion/extension; (b) Forearm supination/pronation; (c) Gyroscope sensor data acquisition for joint range of motion evaluations}
\label{fig9.11}
\end{figure}

The dimensions of the forearm prototype approximate those of a human forearm, with a length of 26 cm and a circumference of 20 cm. Initially, the range of motion of the robotic forearm and elbow is assessed. Fig. \ref{fig9.11} demonstrates the range of motion for elbow flexion/extension and forearm rotation. As listed in Table. \ref{tab2.9}, the elbow flexion/extension exhibits a range of motion from 0 to 145 degrees, while the forearm rotation spans from -60 to 51 degrees. The motion tests are presented in videos 2.1-2.4 in the supplementary materials. The compactness advantages of the robotic arm, demonstrated through the manipulation of objects within limited space, are highlighted in Video 4.1-4.4.

\begin{figure}[htb]
\centerline{\includegraphics[width=0.75\textwidth]{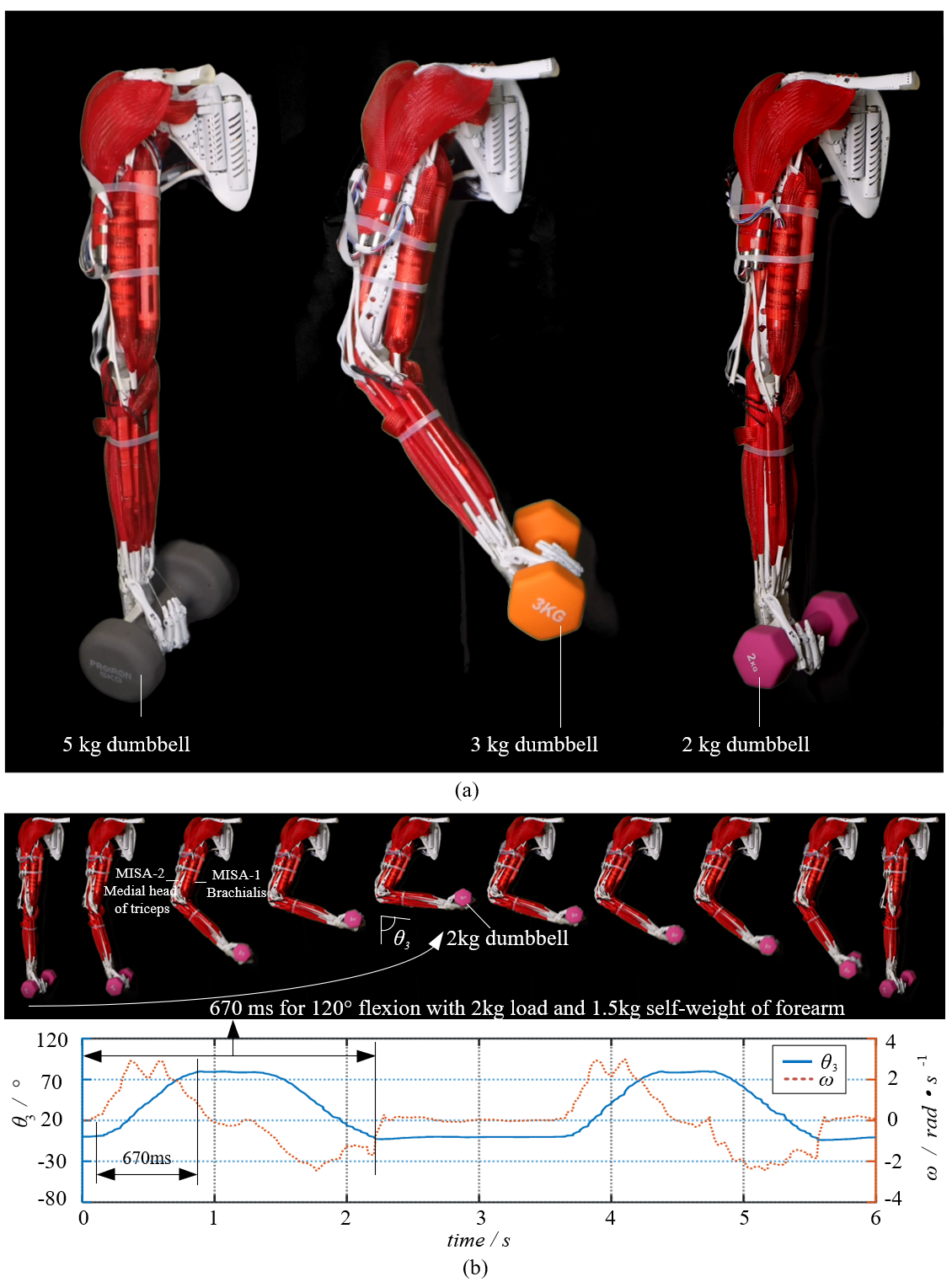}}
\caption{(a) The completed robotic arm prototype holds the dumbbells. (b) The robotic arm lifts the 2 kg dumbbell and the recorded data of forearm position}
\label{fig9.12}
\end{figure}


In order to demonstrate the load capabilities of the biomimetic robotic arm, a non-destructive experiment was conducted. The test involved lifting various weights using the fully assembled arm prototype. As depicted in Fig. \ref{fig9.12}, the robotic arm successfully lifted three different weights, specifically 2kg, 3kg, and 5kg dumbbells. This result showcases the efficient synergy among the MCL, LCL, IOM, annular ligament, and TFCC in maintaining the structural integrity of the elbow and forearm. Notably, no dislocations were observed during the lifting process, and the radius remained stably positioned.

Further testing to ascertain the performance of the biomimetic robotic arm involved a flexion exercise using a 2 kg dumbbell. The elbow was flexed from an extended position to $120^{\circ}$ (Fig. \ref{fig9.12}(b)). The {Maxon motor was equipped with a safety mechanism in the officially provided companion program} designed to curtail any sudden accelerations or excessive speeds. Both the position ($\theta_3$) and angular velocity ($\omega$) of the forearm were documented within the bounds of the maximum permissible speed and acceleration (Fig. \ref{fig9.12}(b)). {The results indicate that the arm achieved full flexion and lifted the 2 kg weight within 0.67 s, achieving a maximum frequency of reciprocal joint movements—defined as the number of complete flexions and extensions accomplished in 1 second—of over 0.74 Hz, excluding intervals of full flexion.} When the angle ($\theta_3$) was set at 50\degree and the angular velocity ($\omega$) at 3 rad/s, {according to calculation,} the joint torque peaked over 12 Nm, inclusive of gravity resistance. Concurrently, the peak power was recorded at 36 W {(calculated from speed and torque)}.

\begin{figure}[htbp]
\centerline{\includegraphics[width=0.75\textwidth]{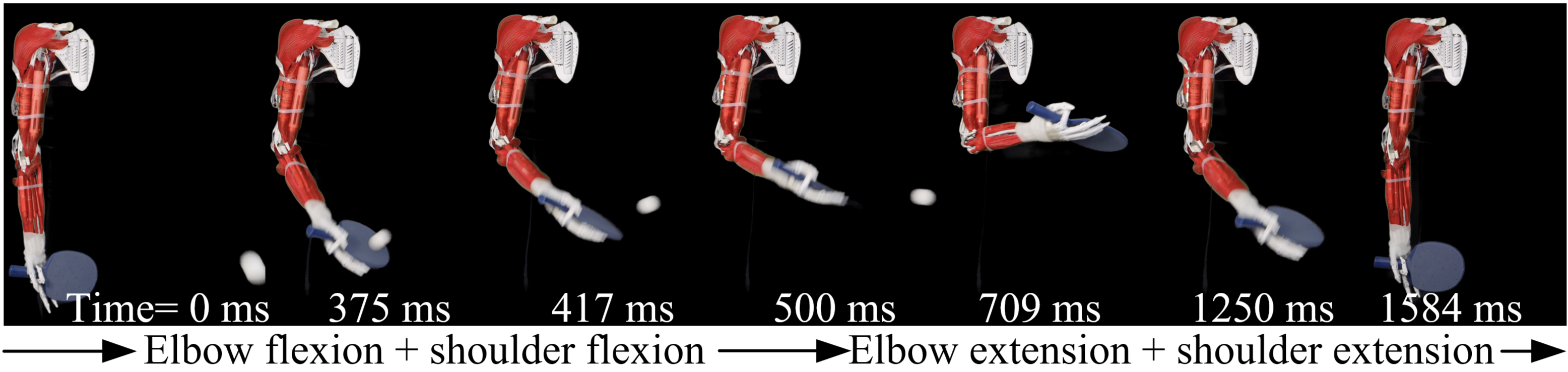}}
\caption{Table tennis playing test.}
\label{fig3.15}
\end{figure}

High-speed performance for a robotic arm's end-effector is an ongoing challenge within the field. A notable example of a high-speed manipulator is Barrett Technology's WAM Arm\cite{senoo2006ball}, with a reported weight of 27 kg and a maximum end-effector speed of 3 m/s. In order to assess high-speed output performance, a table tennis-playing scenario was utilized for the robotic arm under discussion. This scenario involved simultaneous flexing of the elbow and shoulder joints for striking the ping pong ball, followed by a return to the initial position, as illustrated in Fig. \ref{fig3.15}. The trial was performed under minimal load on the compliant actuators, with the ping pong paddle weighing 238 g, thus allowing the actuators to operate at their peak speed of 110 mm/s. {According to calculation,} the end-effector attained an instantaneous speed of 3.2 m/s, with a duration of 188 ms from the onset of arm flexion to the moment of impact with the ping pong ball.

\section{Discussion}

The proposed design, which replicates human biological structures, including bones, ligaments, tendons, and soft actuators with biological muscle performance characteristics, compared to the traditional robotic arm, offers several noteworthy advantages:

Appearance: The prototype is designed to closely mimic the human forearm. Future iterations intend to incorporate artificial skin, further enhancing its resemblance to the human arm both in appearance and structure. Such a humanoid design can foster more intuitive human-robot interactions, reducing the intimidation factor. This is particularly advantageous in settings like healthcare or service sectors where close human-robot collaboration is imperative. A humanoid robotic arm is less likely to be perceived as an unfamiliar entity, promoting wider social acceptance, especially in communal areas. Emulating the human arm not only draws from the biomechanics and movement strategies of humans, informing robot design and control, but also ensures the robot is aptly equipped for tasks designed with human ergonomics in consideration—ranging from door operation to tool usage.

Compactness: The biomimetic forearm structure, in which the radius rotates around the ulna, provides a compact design. With a forearm circumference not exceeding 20 cm, it accommodates over 12 linear actuators for the hand and wrist, each capable of outputting 50 N of force, ensuring dexterity and substantial hand joint output torque.

Safety during Human-Robot Interaction: The system, hinged and fixed by soft tissues {including MCL, LCL, and annular ligament}, resembles a biological body's tension-compression system, exhibiting {passive} damping and flexibility when subjected to external forces. This feature greatly improves safety, as limited external forces can be absorbed by the soft tissues. In cases of excessive external force, the joint can dislocate and recover independently. For irreversible dislocations caused by extreme external forces, manual repairs can be performed without replacing any parts, similar to an orthopaedic doctor repairing a dislocated human joint.

Output Torque: The design achieved a large output torque when compared with the biological joints. Pronation output torque is twice that of biological joint, and supination achieves 85.7\% of its output torque. The entire elbow and forearm have a payload capacity of 4 kg {(The testing was confined to a load of 3kg to preclude any damage to the prototype, foregoing trials under a 4kg load, a limit established through conservative estimation)}, higher than {most} comparable robotic arms (the total weight of the robotic arm is 4 kg, including the shoulder joint), which is listed in Table. \ref{tab4}.

Compared to existing highly biomimetic robotic arms, the proposed design optimize the Load capacity. While conventional robotic arms using hinge joints easily achieve load ability, biomimetic designs with biological joints, such as ECCE\cite{potkonjak2011puller} and Roboy robot\cite{trendel2018cardsflow}, can become unstable when the forearm experiences lateral loads. The inclusion of soft tissues in this design achieves lateral stability akin to hinge joints, resulting in an enhanced load-carrying capacity.

The list of videos for testing the proposed robotic elbow and forearm and demonstrating the capabilities of the robotic arm is provided in Table \ref{tab12}. The supplementary video is accessible via the following link: \url{https://youtu.be/kpVZIUf0f5w}.

\begin{table}[htb]
\caption{Multimedia extensions}
\footnotesize
\begin{center}
\begin{tabular}{l l}
\toprule
No.   & Description  \\
\midrule
Video 1.1 & Improving forearm lateral stability through IOM \\
Video 1.2 & Improving forearm axial stability through IOM \\
Video 1.3 & Variable in MCL strain during elbow movement \\
Video 2.1 & Pronation and supination \\
Video 2.2 & Pronation, supination and wrist flexion\\
Video 2.3 & Elbow flexion/extension with master-slave control \\
Video 2.4 & Elbow flexion/extension \\ 
Video 3.1 & 2kg dumbell lifting test \\ 
Video 3.2 & 3kg dumbell lifting test \\ 
Video 3.3 & Table tennis playing test \\ 
Video 3.4 & Passive performance of the robotic arm \\ 
\makecell[l]{Video 4 \\ \\ \\ \\} & \makecell[l]{Manipulation tests of robotic arm\\including holding a bottle of water; Shaving;\\knocking on the door;\\Putting items on the platform} \\
\bottomrule
\end{tabular}

\begin{tablenotes}
\centering
        \footnotesize
        \item[]  *The video is accessible via the following link: \url{https://youtu.be/kpVZIUf0f5w}
      \end{tablenotes}
      
\label{tab12}
\end{center}
\end{table}

\begin{table*}[htbp]
\caption{{Comparison of different robotic arms}}
\footnotesize
\begin{center}
\begin{tabular}{l l l l l l l l}
\toprule
 Name & Weight (kg)  & Payload (kg)  & Range of motion (\degree)$^A$ & Year & Driven method & Bio$^B$ \\
\midrule

MIA\cite{morita1997development} & 25 & 3 & 0-125,-90-90 & 1997 & Harmonic gear & No \\

Asimo \cite{shigemi2018asimo} & / & 0.5 & / & 2000 & Direct drive & No \\

Hubo 2 \cite{park2007mechanical} & / & 2 & / & 2009 &  Direct drive & No \\

Morgan et al.\cite{quigley2011low} & 11.4 & 2 & / & 2011 & Tendon+Timing Belt & No \\

R1 robot \cite{sureshbabu2017parallel} & / & 1.5 & / & 2017 & Direct drive & No \\

ABB-YuMi \cite{zahavi2018abb} & 9.1 & 0.5 & / & 2017 & Direct drive &  No \\

LIMS\cite{kim2017anthropomorphic} & 5.5 & 2.9 & /& 2017 & Tendon+Timing Belt & No \\

Tsumaki et al.\cite{tsumaki20187} & 2.9 & 1.5 & -90-90,-180-180 & 2018 & Tendon & No \\

Reachy robot\cite{mick2019reachy} & 1.67 & 0.5 & / & 2019 & Direct drive &  No\\

LWH\cite{yang2019lwh} & 3.5 & 0.3 & 0-150,-90-90 & 2019 & Direct drive & No \\

AMBIDEX \cite{choi2020hybrid}& 2.63 & 3 & / & 2020  & Tendon & No \\

P-Rob 2\cite{zhang2021p} & 20 & 3 & -115-115, -162-162 & 2021 & Direct drive & No  \\

Li et al.\cite{li2020modularization} & 2.2 & 1.5 & -130-60, -70-270 & 2021 & Tendon & No \\

Roboy robot\cite{trendel2018cardsflow} & / & / & / & 2013 & Tendon & Yes \\

Kengoro\cite{asano2017design}  & / & / & 0-148, -75-70 & 2017 & Tendon & Yes \\

Kenshiro\cite{potkonjak2011puller}  & / & / & 0-147, n/a  & 2019 & Tendon & Yes \\

ECCE\cite{potkonjak2011puller}  & / & / & / & 2011 & Tendon & Yes \\

\textbf{Proposed deisgn} & 4$^{C}$ & 4 & 0-140.25, -60-51.5 & 2023 & Tendon & Yes \\ 

\bottomrule
\end{tabular}

 \begin{tablenotes}
        \footnotesize
        \item[]  $^A$Range of motion for elbow extension(-)/flexion(+) and forearm pronation(-)/supination(+). $^B$Whether highly biomimetic robotics with biological joints. $^C$Weight of the proposed robotic arm including the shoulder.
      \end{tablenotes}
\label{tab4}
\end{center}
\end{table*}

\section{Conclusion}

In conclusion, this study has developed and validated a novel robotic Elbow-and-Forearm system inspired by the biomechanics of the human skeletal and ligament systems. The research began with a comprehensive investigation of human joint anatomy, highlighting the importance of soft tissues in achieving a balance between compactness, stability, and range of motion. Based on this understanding, a prototype design was proposed, incorporating key soft tissues such as medial collateral ligament, lateral collateral ligament, triangular fibrocartilage complex, annular ligament, and interosseous membrane.

A theoretical analysis of the role of soft tissues in joint stability was conducted, followed by the fabrication of a physical prototype. Through a series of experiments, the proposed skeletal model's resistance to lateral forces and the contribution of soft tissues to stability were assessed. The range of motion and load-carrying capacity of the robotic forearm and elbow were also evaluated, demonstrating the effectiveness of the prototype in replicating human joint capabilities.

Experimental results showed that the range of motion achieved by the robotic forearm and elbow were comparable to human capabilities, and the prototype's ability to lift different dumbbell weights showcased its load-carrying capacity without dislocation or significant displacement. This research not only contributes to a better understanding of human arm biomechanics but also advances the development of more sophisticated robotic prosthetics and exoskeletons. The findings have the potential to pave the way for further innovation in the field.

\newpage

\bibliographystyle{./IEEEtran} 
\bibliography{./IEEEabrv,./IEEEexample}

\end{document}